\documentclass{article}

\usepackage[preprint]{neurips_2025}

\usepackage[utf8]{inputenc} 
\usepackage[T1]{fontenc}    
\usepackage{hyperref}       
\usepackage{url}            
\usepackage{booktabs}       
\usepackage{amsfonts}       
\usepackage{nicefrac}       
\usepackage{microtype}      
\usepackage{xcolor}         
\usepackage{graphicx}       
\usepackage{enumitem}       
\usepackage{bm}             
\usepackage{xspace}
\usepackage{amsmath}
\usepackage{wrapfig}
\usepackage{multirow}
\usepackage[most]{tcolorbox}
\usepackage{colortbl}
\usepackage{arydshln}
\usepackage{subfigure}
\usepackage{float}
\usepackage{listings}
\usepackage{comment}

\lstdefinestyle{python}{
    backgroundcolor=\color[RGB]{250,250,250},   
    commentstyle=\itshape\color[RGB]{172,172,189},
    keywordstyle=\color[RGB]{78,147,206},
    numberstyle=\fontfamily{fvm}\selectfont\tiny\color[RGB]{172,172,189},
    stringstyle=\color[RGB]{74,163,128},
    basicstyle=\fontfamily{fvm}\selectfont\small,
    breakatwhitespace=false,         
    breaklines=true,                 
    captionpos=t,                    
    keepspaces=true,                 
    numbers=left,                    
    numbersep=10pt,                  
    showspaces=false,                
    showstringspaces=false,
    showtabs=false,                  
    tabsize=4,
    language=Python,
    frame=single,
    rulecolor=\color{white},
    framexleftmargin=2em,
    framesep=5pt,
    framerule=0pt,
    literate=
    *{True}{{{\color[RGB]{78,147,206}True}}}{4}
     {False}{{{\color[RGB]{78,147,206}False}}}{5}
     {...}{{{\color[RGB]{205,65,120}...}}}{3}
     {0}{{{\color[RGB]{205,65,120}0}}}{1}
     {1}{{{\color[RGB]{205,65,120}1}}}{1}
     {2}{{{\color[RGB]{205,65,120}2}}}{1}
     {3}{{{\color[RGB]{205,65,120}3}}}{1}
     {4}{{{\color[RGB]{205,65,120}4}}}{1}
     {5}{{{\color[RGB]{205,65,120}5}}}{1}
     {6}{{{\color[RGB]{205,65,120}6}}}{1}
     {7}{{{\color[RGB]{205,65,120}7}}}{1}
     {8}{{{\color[RGB]{205,65,120}8}}}{1}
     {9}{{{\color[RGB]{205,65,120}9}}}{1},
}

\newcommand{\onezero}{1\xspace}
\newcommand{\twozero}{2\xspace}
\newcommand{\name}{KumoRFM\xspace}
\newcommand{\nameone}{\name-\onezero}
\newcommand{\nametwo}{\name-\twozero}

\definecolor{kumo1}{RGB}{238,118,151}
\definecolor{kumo2}{RGB}{244,179,197}
\definecolor{kumo3}{RGB}{249,219,227}
\definecolor{kumoyellow}{RGB}{239,214,89}

\definecolor{color1}{RGB}{254,191,185}
\definecolor{color2}{RGB}{254,128,41}
\definecolor{color3}{RGB}{254,15,127}
\definecolor{color4}{RGB}{10,153,201}
\definecolor{color5}{RGB}{5,100,18}

\definecolor{purple}{RGB}{147,7,204}
\definecolor{purple2}{RGB}{161,67,233}
\definecolor{green}{RGB}{159,213,179}
\definecolor{blue}{RGB}{10,153,201}
\definecolor{red}{RGB}{213,42,45}
\definecolor{orange}{RGB}{254,128,41}
\definecolor{gray}{RGB}{120,120,120}
\definecolor{pink}{RGB}{207,14,133}
\definecolor{macred}{RGB}{237,106,94}
\definecolor{macyellow}{RGB}{244,191,79}
\definecolor{macgreen}{RGB}{97,197,84}

\newcommand{\cellfirst}{\cellcolor{green!50!white}}
\newcommand{\cellsecond}{\cellcolor{green!25!white}}

\tcbset{
  explanation/.style={
    enhanced,
    frame hidden,
    arc=10pt,
    boxsep=5pt,
    interior style={left color=black!5!white,right color=white},
  }
}

\title{
\nametwo{}: Scaling Foundation Models\\for Relational Learning
}

\begin{document}

\maketitle

\begin{abstract}

We introduce \emph{\nametwo}, the next iteration of a pre-trained foundation model for relational data. \nametwo supports in-context learning as well as fine-tuning and is applicable to a wide range of predictive tasks.
In contrast to tabular foundation models, \nametwo natively operates on relational data, processing one or more connected tables simultaneously without manual table flattening or target variable generation, all while preserving temporal consistency.

\nametwo leverages a large corpus of synthetic and real-world data to pre-train across four axes: the row and column dimensions at the individual table level, and the foreign key and cross-sample dimensions at the database level.
In contrast to its predecessor, \nametwo injects task information as early as possible, enabling sharper selection of task-relevant columns and improved robustness to noisy data.

Through extensive experiments on 41 challenging benchmarks and analysis around expressivity and sensitivity, we demonstrate that \nametwo outperforms supervised and foundational approaches by up to 8\%, while maintaining strong performance under extreme settings of cold start and noisy data.
To our knowledge, this is the first time a few-shot foundation model has been shown to surpass supervised approaches on common benchmark tasks, with performance further improving upon fine-tuning.
Finally, while \nameone was limited to small-scale in-memory datasets, \nametwo scales to billion-scale relational datasets.

\end{abstract}

\section{Introduction}
\label{sec:introduction}

Foundation models are fundamentally changing the landscape of predictive modeling on structured data, achieving state of the art accuracy~\citep{tabicl,tabpfn, contexttab, garg2025realtabpfn, limix, fey2025kumorfm, qu2026tabiclv2}.
The key enabling factor for this paradigm shift is the rise of \emph{In-Context Learning (ICL)}~\citep{gpt3}, which enables solving novel predictive tasks in a single feed-forward step, avoiding expensive task-specific training.

Traditional supervised machine learning methods require training on large datasets containing labeled data for a single specific task. In contrast, ICL allows for \emph{pre-training} a general model on large-scale synthetic and real data, which can adapt to novel, unseen tasks by showing it a few labeled examples — the \emph{context} — at prediction time only.
The consequences are dramatic: data requirements for solving unseen predictive tasks are substantially reduced.
Similarly, time to deployment for novel tasks is reduced from several hours and days to seconds.

ICL was originally introduced in the context of \emph{Large Language Models (LLMs)}~\cite{transformer, gpt2, gpt3}, where models are conditioned on a sequence of input/output examples at inference time.
Despite their strong generalization capabilities, the language-based inductive bias of LLMs is not well aligned with predictive tasks over structured data (\emph{cf.}~Sec.~\ref{sec:results}).
Therefore, \emph{Tabular Foundation Models (TFMs)} extended the principle, allowing for in-context learning on data populating a single table~\citep{tabpfn}.
They can be naively extended to the relational setting by flattening multiple tables via fixed-function, parameter-free encoders~\citep{hayler2025bringing,eremeev2025turning,
xu2026rdblearn, wang2026reliclsynthetic}.

However, these approaches remain limited for real-world applications: they rely on flattening relational data into single, row-wise representations.
This either necessitates substantial feature engineering and target label generation (undermining their key advantage of training-free, ad-hoc predictions), or relies on generic, task-agnostic feature aggregation schemes that fail to exploit task-specific relational structure for downstream prediction. Enabling foundation models to operate directly on multi-table relational data would eliminate these limitations, allowing truly end-to-end, training-free predictions while fully leveraging the rich structure inherent in real-world datasets.

Here, we introduce \emph{\nametwo}, a foundation model for general data and tasks that live in relational databases, \emph{e.g.}, predicting user churn, inventory demand, or detecting fraud.
\nametwo is the next iteration of a ``database-native'' \emph{Relational Foundation Model (RFM)} \citep{fey2025kumorfm}.
We show that pre-training on relational data enables the model to effectively learn how to filter, correlate, and aggregate information across multiple tables.
Compared to v\onezero, \nametwo performs ICL over two separate attention stages, one for intra-table processing and one for aggregating information across tables.
This design enables task-conditioned feature extraction at both the table and database level, leading to improved predictive performance.

In comprehensive experiments across 41 predictive tasks on real-world databases from RelBenchV1~\citep{relbench}, RelBenchV2~\citep{gu2026relbenchv2}, SALT~\citep{salt} and 4DBInfer~\citep{wang2024dbinfer}, we demonstrate that \nametwo outperforms both supervised models as well as few-shot foundation models across tabular and relational settings.
Specifically, on RelBenchV1, \nametwo outperforms v\onezero by $\approx$10\% and even surpasses its closest competitor, the task-specific supervised RelGNN~\citep{relgnn}, by $\approx$5\% for both classification and regression tasks.
Furthermore, on the SAP SALT enterprise benchmark, \nametwo achieves state-of-the-art performance, surpassing giant tabular AutoGluon ensembles~\citep{autogluon} by $\approx$8\% and recent tabular foundation models~\citep{qu2026tabiclv2} by $\approx$25\%.
Performance can be further improved by $\approx$16\% through fine-tuning.
This highlights the benefits of modeling relational structure directly rather than relying on flattened tabular representations.
Finally, we analyze the robustness of \nametwo with respect to context size, subgraph size, feature and link scarcity, and feature noise.
In general, we observe that \nametwo is able to preserve strong predictive capability even under extreme settings of cold start, as well as missing or incomplete data.

Overall, the results suggest that the relational domain is undergoing the same paradigm shift already witnessed in other domains, where few-shot foundation models are now able to rival carefully tuned, task-specific approaches.
This, in turn, enables accurate on-demand predictions on relational data that can be seamlessly integrated into agentic workflows for real-time decision-making.

\nametwo is publicly available through our \href{https://kumorfm.ai}{Python SDK}.
While previously limited to in-memory datasets, \nametwo supports direct connectivity to SQL databases and cloud data warehouses, enabling it to scale seamlessly to databases of arbitrary size.
Specifically, data processing can either be pushed down to the underlying database or executed via an SSD-based graph engine that leverages memory-mapped I/O to enable high-throughput, low-latency access to databases with 500B+ rows.

\section{Relational Foundation Models and \name}
\label{sec:rfm}

\name is a pre-trained relational foundation model~\citep{fey2025kumorfm}, combining the concepts of \emph{Relational Deep Learning}~\citep{rdl} and in-context learning~\citep{gpt3}:

\paragraph{Relational Deep Learning (RDL).}
RDL models~\citep{rdl} a relational database as a \emph{temporal heterogeneous graph} \mbox{$\mathcal{G} = (\mathcal{V}, \mathcal{E})$}, where each record in any table corresponds to a node $v \in \mathcal{V}$, and the primary-foreign key links define the edges $(v, w) \in \mathcal{E}$.
Every node is associated with a timestamp $t$, allowing us to define a time-consistent snapshot $\mathcal{G}^{\le t}$ that contains only the records available up to that moment.
We define a \emph{training example} $(v_i, t_i, y_i)$ for a specific node $v_i$ at a specific point in time $t_i$, where $y_i$ is the ground-truth target. A sequence of examples $\mathcal{T} = \{(v_i, t_i, y_i)\}_{i=1}^{n-1}$, called \emph{task table}, implicitly defines a prediction task, used as training set for supervised learning.

In single tabular learning, learning can be simply performed over rows in $\mathcal{T}$~\citep{tabpfn, autogluon}.
The relational setting comes with additional challenges.
Relevant input features are not given in the task table itself but live in multiple tables, connected to $v_i$ via complex relational structures.
Thus, for each $v_i$, task-relevant signals need to be extracted from related data sources, \emph{i.e.} within local subgraphs $\mathcal{G}^{\leq t_i}[v_i]$ around $v_i$ up to timestamp $t_i$.
Recent work has made strong progress in processing such data through advances in model architectures~\citep{relgt}, graph rewiring techniques~\citep{relgnn}, and pre-training strategies~\citep{ranjan2026rt}.

\paragraph{Relational Foundation Models (RFMs).}

RFMs enable predictive modeling on any relational database without requiring task-specific training~\citep{fey2025kumorfm,wang2026reliclsynthetic, plurel, xu2026rdblearn,ranjan2026rt}.
Specifically, RFMs perform few-shot predictions by treating training examples as \emph{context examples} for in-context learning~\citep{fey2025kumorfm}.
That is, a single feed-forward step extracts signal from subgraphs within the context to inform predictions.
More formally, the task table $\mathcal{T} = \{(v_i, t_i, y_i)\}_{i=1}^{n-1}$, containing the context examples, defines what information to use to predict the label $\hat{y}_n$ for a node $v_n$ for a future point in time $t_n$:
\[
\hat{y}_n = \mathrm{RFM}^{\includegraphics[height=8pt,keepaspectratio]{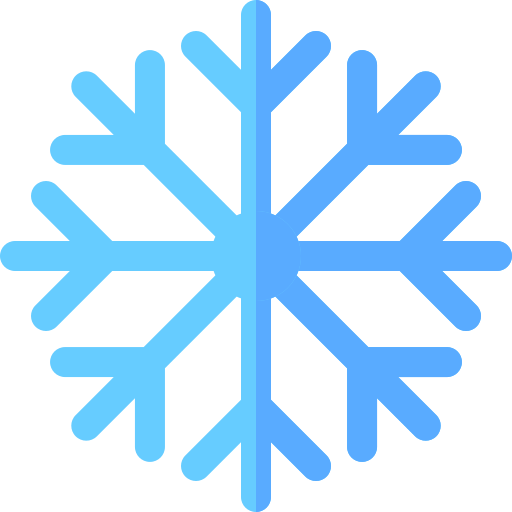}}_{\bm{\theta}}(\mathcal{G}^{\leq t_n}[v_n], \{(\mathcal{G}^{\leq t_i}[v_i],y_i)\}_{i=1}^{n-1}) \textnormal{.}
\]

Because the RFM parameters $\bm{\theta}$ are pre-trained and remain frozen during this process, the model must entirely rely on its ability to reason over the relational structure of input subgraphs $\mathcal{G}^{\leq t_i}[v_i]$ and outputs $y_i$ to generalize to the specified task for the new example.
This ability naturally extends the paradigm of tabular foundation models to the relational multi-table and temporal setting, building on recent advancements in synthetic data generation~\citep{plurel,wang2026reliclsynthetic,hoppe2025generating,tabpfn,mitra,breejen2025tabforest,qu2026tabiclv2}, multi-scale processing~\citep{orion}, or semantic awareness~\citep{contexttab}.

\paragraph{The KumoRFM System.}

\begin{figure}[t]
\centering
\includegraphics[width=\linewidth]{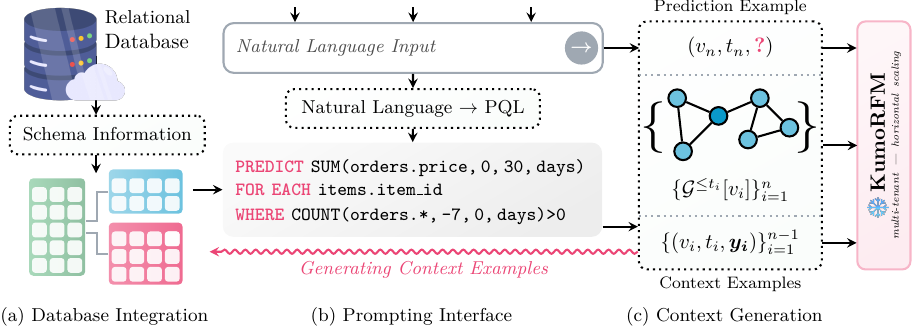}
\caption{\textbf{Overview of the \name system.}
\textbf{(a)} \name operates directly on a relational database, where the schema defines the underlying graph structure.
\textbf{(b)} Users specify predictive tasks via a declarative Predictive Query Language or natural language.
\textbf{(c)} \name constructs context examples $(v_i, t_i, y_i)$ and corresponding input subgraphs $\mathcal{G}^{\leq t_i}[v_i]$, where targets $y_i$ are derived according to the query definition.
All data processing is pushed down to the underlying database for scalable execution, after which the constructed context is passed to the model for inference.
}
\label{fig:end2end}
\end{figure}

\name (\emph{cf.}~Fig.~\ref{fig:end2end}) is an end-to-end system built around a Relational Foundation Model for predictive tasks on structured data.
Beyond the model itself, \name orchestrates a set of tightly coupled components to enable low-latency predictions and seamless integration into agentic workflows.
In particular, \name manages the following components:
\begin{itemize}[label=\textcolor{kumo1}{\textbullet},leftmargin=10pt]
\item \textbf{Database integration (Fig.~\ref{fig:end2end}(a)):}
\name integrates natively with relational databases, allowing it to construct graphs and feature spaces directly from database metadata or a semantic model.
This integration is a one-time setup that relies primarily on schema-level information, making it fast and lightweight in practice.
Support for views and derived columns enables flexible graph rewiring without requiring costly data materialization or preprocessing pipelines.
Crucially, this integration allows \name to scale directly with the underlying database, regardless of its size.
\item \textbf{Prompting interface (Fig.~\ref{fig:end2end}(b)):}
Interaction with \name is facilitated through the high-level declarative \emph{Predictive Query Language (PQL)}~\citep{kocijan2026pql}.
Queries are expressed as compositions of aggregations, filters, and binary operations, providing a concise and flexible abstraction for specifying predictive tasks.
This interface supports a wide range of use cases, from basic tasks such as value imputation to more complex temporal prediction problems (\emph{e.g.}, user churn and item demand forecasting).
It also serves as a structured intermediate representation for natural language interfaces, avoiding the need for agents to construct model inputs directly.
As such, PQL turns predictive modeling into a composable primitive for agents, allowing queries to be stacked and combined with SQL operations to express complex decision-making pipelines.
\end{itemize}

\vspace{-0.8cm}

\begin{wrapfigure}{r}{0.44\textwidth}
  \centering
  \vspace{0.435cm}
  \includegraphics[width=0.43\textwidth]{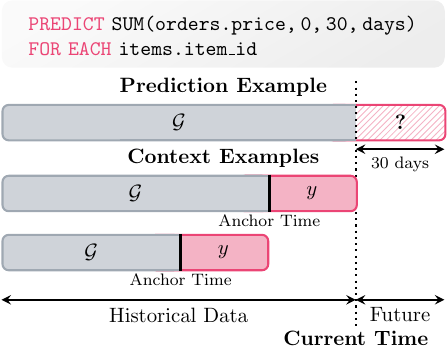}
  \caption{\textbf{Automatic context generation.} The query~\citep{kocijan2026pql} defines how context input/output pairs can be automatically constructed from historical database states while preventing data leakage.}\vspace{-0.5cm}
  \label{fig:autocontext}
\end{wrapfigure}

\noindent
\begin{itemize}[label=\textcolor{kumo1}{\textbullet},leftmargin=10pt]
\item \textbf{Automatic context generation (Fig.~\ref{fig:end2end}(c)):}
Given a predictive query and graph metadata, the full context can be constructed ad-hoc by pushing computation to the underlying data backend. In particular, ground-truth labels $\{(v_i, t_i, y_i)\}_{i=1}^{n-1}$ can be obtained in two ways: either directly from existing database values (\emph{e.g.}, in value imputation settings), or by computing them at earlier timestamps according to the query definition.
Similarly, input subgraphs $\{ \mathcal{G}^{\leq t_i}[v_i] \}_{i=1}^{n}$ are constructed by traversing the database schema along valid relational paths.

As such, \name supports both static as well as temporal predictive tasks. For static tasks (\emph{e.g.}, ``predict user age''), the target for context examples can be generated by sampling users whose age is already known. For temporal tasks (\emph{e.g.}, ``forecast sales next quarter'' or ``predict transaction fraud probability at time $t_i$''), the context is effectively generated by ``replaying'' historical states of the database, sampling entity-centric subgraphs up to anchor time $t_i$ and computing the corresponding target $y_i$ based on events occurring after $t_i$ (\emph{cf}.~Fig.~\ref{fig:autocontext}).
In practice, only a small number of context examples (\emph{e.g.}, up to $10k$) is sufficient to provide strong signal for in-context learning while preserving low latency.
\end{itemize}

\vspace{-0.3cm}
\begin{itemize}[label=\textcolor{kumo1}{\textbullet},leftmargin=10pt]
\item \textbf{Scalability:} \nametwo{} is designed to support in-context learning natively over databases, either \textbf{(1)} in-memory, \textbf{(2)} via database and warehouse connectors, or \textbf{(3)} through a custom graph engine, optimized for fast ingestion and neighbor retrieval.
The in-memory mode is ideal for exploration and small-scale experimentation, while database connectors and graph engine are designed for production workloads at scale.
In the database connector mode, no intermediate graph representation is constructed.
Instead, both context retrieval and subgraph construction are \emph{pushed down} to the database via recursive SQL execution along metapaths.
The graph engine builds an intermediate memory-mapped data structure, enabling \nametwo to scale to 500B+ rows with high throughput (up to 5GB/sec bandwidth, 20M lookups/sec) and low-latency requirements.
\end{itemize}

\begin{itemize}[label=\textcolor{kumo1}{\textbullet},leftmargin=10pt]
\item \textbf{Stateless, multi-tenant model serving:}
Kumo's relational foundation model can be deployed as a stateless inference service, decoupled from any specific database, task, or tenant.
All required context is provided at request time, enabling strict isolation across workloads and eliminating the need for persistent state or per-tenant model instances.
This design simplifies scaling, supports efficient multi-tenant scheduling, and allows horizontal replication without coordination.
Inputs are processed ephemerally and are not retained beyond the scope of a request.
\end{itemize}
\begin{itemize}[label=\textcolor{kumo1}{\textbullet},leftmargin=10pt]
\item \textbf{Predictions, embeddings, evaluations and explanations:}
Beyond producing predictive outputs, \name returns entity embeddings for downstream use and built-in evaluation metrics to quantify predictive performance.
Furthermore, the system incorporates mechanisms for explainability, offering insights into the reasoning processes that lead to its predictions.
\end{itemize}

Taken together, these components enable \name to function as an agent-ready system.
To facilitate integration, we provide a portable collection of skills\footnote{\name skills: \url{https://github.com/kumo-ai/kumo-coding-agent}} for modern LLM-based tooling.

\section{The \nametwo{} Architecture}
\label{sec:kumorfmv2}

\begin{figure}[t]
\centering
\includegraphics[width=\linewidth]{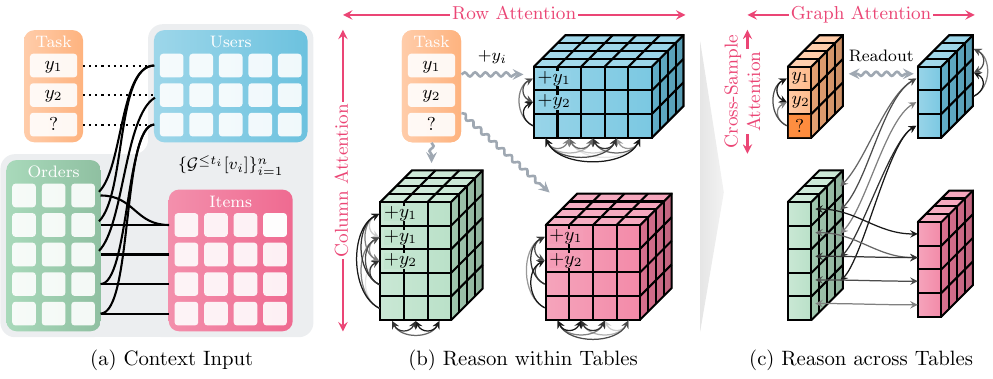}
\caption{\textbf{Overview of the \nametwo architecture.}
\textbf{(a)} Given a task table with connected tables as context examples, \textbf{(b)} \nametwo first computes representations for all rows in the context via alternating column- and row-wise attention.
These representations are made \emph{task-conditioned} by incorporating context targets directly into the input tables.
\textbf{(c)} The representations are then enriched via graph attention over primary–foreign key relationships and cross-sample attention over context examples, enabling in-context learning beyond single tables. Finally, the target is predicted from the resulting representation of the prediction example.
}
\label{fig:overview}
\end{figure}

\nametwo follows the general blueprint of RFMs as introduced in \nameone~\citep{fey2025kumorfm} as well as in Sec.~\ref{sec:rfm}. A schematic overview of the architecture is shown in Fig.~\ref{fig:overview}.
The architecture is highly general, allowing it to process arbitrary relational databases and to make predictions in a wide range of formats. It 
is trained on a large distribution of tabular and relational tasks, enabling it to efficiently reason over diverse structural priors and capture complex non-linear dependencies between inputs and outputs.
This allows the model to generalize effectively to unseen databases and tasks during inference, beyond those encountered during training.

\nametwo powers a \emph{database-native} design to simultaneously master three distinct scales of information:
It reasons over input context examples at the \textbf{(1)} \emph{row/column level} within individual tables to understand and relate specific multi-modal attributes, at the \textbf{(2)} \emph{foreign-key level} to capture the structural topology and relationships across tables, and at the \textbf{(3)} \emph{cross-sample level} to recognize and share patterns across context examples. Context targets are provisioned to the process early, allowing the model to perform \emph{task-conditioned} processing throughout the model.
\nametwo represents a significant improvement over \nameone.
Below, we summarize its key changes:

\paragraph{Hierarchical Attention Scheme.}

Optimizing performance across these various scales requires hierarchical attention at the table and graph levels.
A lightweight network first extracts task-conditioned row embeddings from individual tables through alternating column and row attention. 
A larger network then distributes and relates these embeddings across tables and context samples via foreign key and cross-sample attention. 
This staged hierarchical design produces better cell- and structure-aware representations while enabling the removal of noisy data early on.
It also avoids the quadratic complexity of attending to each table cell.
Instead, the attention scheme scales to large context sizes while preserving the ability to attend across row, column, foreign key and cross-sample dimensions.

\paragraph{Smarter Context and Lagged Targets.}

Running foundation models over millions of context examples remains an open research question, making effective context selection critical at scale.
\nametwo addresses this with a structured context selection strategy that combines \emph{local} context drawn from the prediction entity itself with \emph{global} context from the most recent database snapshots.
This allows the model to condition on its own lagged targets and prior subgraphs while incorporating up-to-date global information from other entities in the database.

\paragraph{Training Data.}
\nametwo{} is pre-trained on an expanded combination of synthetic and real-world data. 
Synthetic tables and graphs are generated via \emph{Structural Causal Models}~\citep{pearl2009causality, peters2017elements,tabpfn,hoppe2025generating,plurel},
while real-world pre-training data consists of publicly available relational databases paired with a diverse set of predictive queries, executed over a set of random entities at different points in time.
Pre-training is conducted in multiple stages, transitioning from single tables to more complex relational structures.

\paragraph{Ensembling.}

In contrast to v\onezero, \nametwo enables more fine-grained control over ensembling. This includes randomized column and class order shuffles to reduce sensitivity to input permutations, as well as ensembling over the number of hops and various post-processing techniques.
Together, these mechanisms reduce variance across runs and improve robustness to arbitrary input orderings.

\paragraph{Expressivity.}
\label{sec:counterexample}

Recent approaches apply tabular foundation models to multi-table relational tasks via propositionalization~\citep{kramer2001propositionalization,kanter2015deep}, which pre-aggregates relational subgraphs into a single flat table~\citep{hayler2025bringing,eremeev2025turning,
xu2026rdblearn, wang2026reliclsynthetic,gnntabpfn}.
We argue that such task-agnostic, fixed-function pipeline induces a fundamental expressivity gap.
For example, the \emph{deep feature synthesis (DFS)}~\citep{kanter2015deep} utilized in \citet{xu2026rdblearn} and in \citet{wang2026reliclsynthetic} applies column-wise pre-defined aggregations, thereby limited to capture interactions across rows.

\begin{figure}[t]
\centering
\includegraphics[width=\linewidth]{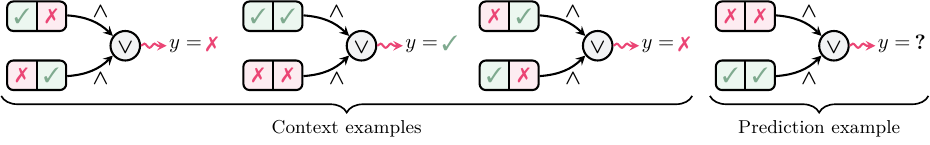}
\caption{\textbf{An adversarial example for fixed-function column-wise encoders.} The label is positive iff both child features co-occur within at least one connected row.
Row-level alignment is required to determine the label, which column-wise encoders~\citep{kanter2015deep} fail to capture.
}
\label{fig:conjunction}
\end{figure}

To illustrate this, we consider a database with a primary table $T_1$ and a child table $T_2$ containing binary features $A$ and $B$.
For each entity in $T_1$, we define the target $y=1$ iff $A$ and $B$ co-occur in at least one related row of $T_2$ (\emph{cf}.~Fig.~\ref{fig:conjunction}).
Consequently, any model on top of column-wise encoders \citep{xu2026rdblearn, wang2026reliclsynthetic} fails to distinguish the two classes due to uninformative column-wise marginals (\textbf{AUROC = 0.5}), whereas \nametwo with its task-conditioned feature extraction scheme achieves perfect separation (\textbf{AUROC = 1.0}).

\paragraph{Fine-Tuning.}

\nametwo can be fine-tuned and specialized to a specific dataset and task to achieve maximum performance.
In this setting, the model weights now encode domain- and task-specific information, resulting in richer parametric knowledge.
Consequently, fine-tuning reduces reliance on in-context learning and context selection.
This is particularly beneficial in large-scale regimes, where the available training data far exceeds current context size limitations.

\section{Experimental Results}
\label{sec:results}

For the experimental evaluation, we assess the in-context learning capabilities of \nametwo{} on four different benchmark suites: RelBenchV1~\citep{relbench}, RelBenchV2~\citep{gu2026relbenchv2}, 
SALT~\citep{salt}, and 4DBInfer~\citep{wang2024dbinfer}.
None of these datasets were used during pre-training, which guarantees no leakage of information.
We consider binary classification (Sec.~\ref{subsec:binary_classification}), multi-class classification (Sec.~\ref{subsec:multi_classification}), and regression tasks (Sec.~\ref{subsec:regression}).
Besides raw performance, we analyze the robustness of \nametwo across varying data regimes, specifically analyzing its performance under conditions of extreme data scarcity, high feature sparsity, and structural noise (Sec.~\ref{subsec:ablations}).
Finally, we analyze the impact of fine-tuning on model performance (Sec.~\ref{subsec:finetuning}).

\begin{figure}[t]
\centering
\includegraphics[width=\linewidth]{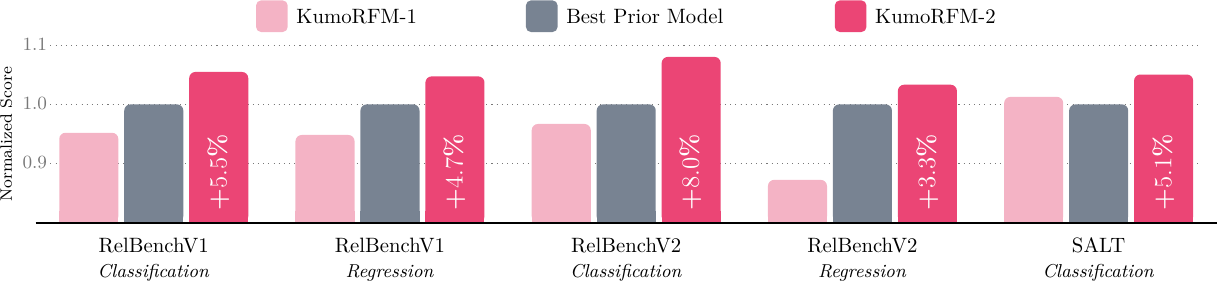}
\caption{\textbf{Performance overview of \nametwo}, where scores are normalized relative to the strongest (supervised) model on each benchmark suite.
\nametwo shows a significant improvement over \nameone and consistently outperforms prior supervised or foundational few-shot approaches. Further details are provided in 
Tables~\ref{tab:rbv1-cls}, \ref{tab:rbv2-cls}, \ref{tab:salt}, \ref{tab:dbinfer}, \ref{tab:rbv1-reg}, and \ref{tab:rbv2-reg}.}
\label{fig:bar}
\end{figure}

We closely follow the evaluation protocols of the utilized benchmarks, ensuring that \nametwo adheres to the same temporal constraints and data splits as prior models.
Since \nametwo does not require task-specific training, we utilize data from both the training and validation splits (if provided) to populate the model's context.
We use at most $10k$ context examples to reflect realistic settings and ensure fair comparison with prior work.
Unless otherwise stated, no further training or fine-tuning of \nametwo is performed, and all predictions are obtained solely through in-context learning on the base model.
Benchmark scripts for all experiments are publicly available\footnote{Benchmark scripts: \url{https://github.com/kumo-ai/kumo-rfm}}.
An overview of \nametwo's performance across the different benchmarks is shown in Fig.~\ref{fig:bar}. 

\paragraph{Databases and Tasks.}

\begin{table}
\centering
\caption{\textbf{Database summary and statistics across RelBenchV1, RelBenchV2, SALT and 4DBInfer.}
Datasets vary in the number of tables, rows, columns, and available training examples (\#Train).
ICL Coverage denotes the fraction of training data used for in-context learning, reaching as low as 0.2\%.
}
\label{tab:datasets}
\setlength{\tabcolsep}{3pt}
\resizebox{\textwidth}{!}{
\begin{tabular}{llllrrrrrr}
\toprule
& \multirow{2}{*}{\textbf{Dataset}} & \multirow{2}{*}{\textbf{Domain}} & \multicolumn{2}{l}{\multirow{2}{*}{\textbf{Tasks}~~~~~~~~~~~~~~~~~~~~~\textbf{\#Tasks}}} & \multicolumn{3}{l}{\multirow{2}{*}{\textbf{\#Tables}~~\textbf{\#Rows}~~\textbf{\#Cols}}} & \multirow{2}{*}{\textbf{\#Train}} & \multicolumn{1}{c}{\textbf{ICL}} \\
& & & & & & & & & \multicolumn{1}{c}{\textbf{Coverage}} \\
\midrule
\multirow{7}{*}{\rotatebox{90}{\tiny \textsc{\textcolor{gray!70!black}{RelBenchV1~~}}}}
& \texttt{f1} & Sports & 
Race outcome & 3 & 9 & 74,063 & 67 & $\leq$11,977 & 83.5\% \\
& \texttt{avito} & E-commerce & User engagement, CTR & 3 & 8 & 20,679,117 & 42 & $\leq$116,598 & 8.6\% \\
& \texttt{event} & Social & Event participation & 3 & 5 & 41,328,337 & 128 & $\leq$23,424 & 42.7\% \\
& \texttt{trial} & Medical & Clinical trial outcome & 3 & 15 & 5,434,924 & 140 & $\leq$12,954 & 77.2\% \\
& \texttt{amazon} & E-commerce & Churn, LTV & 4 & 3 & 15,000,713 & 15 & $\leq$5,142,347 & 0.2\% \\
& \texttt{stack} & Social & Community activities & 3 & 7 & 4,247,264 & 52 & $\leq$3,633,674 & 0.3\% \\
& \texttt{hm} & E-commerce & Churn, LTV & 2 & 3 & 16,664,809 & 37 & $\leq$3,947,966 & 0.3\% \\
\cdashline{2-10}
\multirow{3}{*}{\rotatebox{90}{\tiny \textsc{\textcolor{gray!70!black}{RelBenchV2~}}}}
& \texttt{mimic} & Medical & Length-of-stay & 1 & 6 & 2,424,751 & 54 & 16,999 & 58.8\% \\
& \texttt{ratebeer} & E-commerce & Engagement, Churn & 4 & 13 & 13,787,005 & 221 & $\leq$2,563,053 & 0.4\%  \\
& \texttt{arxiv} & Academic & Citations, Publications & 2 & 6 & 2,146,112 & 21 & $\leq$249,784 & 4.0\% \\
\cdashline{2-10}
& SALT & ERP & Business outcome & 8 & 4 & 4,257,145 & 31 & $\leq$1,916,610 & 0.5\% \\
\cdashline{2-10}
\multirow{4}{*}{\rotatebox{90}{\tiny \textsc{\textcolor{gray!70!black}{4DBInfer~~}}}}
& AB & E-commerce & Churn & 1 & 3 & 24,291,489 & 15 & 1,194,773 & 0.8\% \\
& OB & Social & CTR & 1 & 8 & 2,170,441,217 & 31 & 78,424 & 12.4\% \\
& RR & E-commerce & CVR & 1 & 3 & 23,033,676 & 11 & 90,003 & 11.1\% \\
& SE & Social & Churn, Popularity & 2 & 7 & 5,399,818 & 49 & $\leq$347,285 & 2.9\% \\
\midrule
& \textbf{Total} & & & 41 & 100 & 2,349,210,440 & 914 & $\leq$68,020,630 & 0.6\% \\
\bottomrule
\end{tabular}
}
\end{table}

We evaluate \nametwo on a diverse collection of benchmark suites comprising 15 relational databases and 41 temporal prediction tasks.
These benchmarks span a wide range of domains, including e-commerce, Q\&A platforms, medical, academic, enterprise resource planning (ERP), and sports.
The tasks cover a variety of temporal prediction problems, such as churn, lifetime value (LTV), click-through rate (CTR), and conversion rate (CVR).

Databases are summarized in Table~\ref{tab:datasets}, which vary significantly in the numbers of tables, rows, columns, and available training examples.
We also report the coverage of utilized training examples $\frac{\textrm{\#context examples}}{\textrm{\#available training examples}}$ for in-context learning in foundation models .
Notably, in the largest settings, only up to 0.2\% of available training data is actually being used as context examples.

Specifically, we evaluate on all binary classification and regression tasks in RelBenchV1~\citep{relbench}, the standard benchmark for relational deep learning, and its recent extension RelBenchV2~\citep{gu2026relbenchv2}.
In addition, we include the enterprise-scale SALT database~\citep{salt}, which captures real-world customer interactions in ERP systems, where common missing fields in business processes must be predicted.
ERP systems are critical for managing core business operations, including finance, human resources, production, and supply chains.
SALT includes 8 large multi-class classification tasks, with up to 589 classes in the most challenging setting.
Finally, we evaluate on binary classification tasks in 4DBInfer~\citep{wang2024dbinfer}, which extends RelBench with additional databases and tasks, particularly for relationship attribute prediction.

\paragraph{Baselines.}

\begin{table}
\centering
\caption{\textbf{Overview of utilized baselines}.
We compare against supervised tabular models (including hand-crafted and automatic feature engineering pipelines for flattening multi-table data), state-of-the-art relational deep learning models, and language-based, tabular, and relational foundation models.}
\label{tab:baselines}
\setlength{\tabcolsep}{3pt}
\resizebox{\textwidth}{!}{
\begin{tabular}{lrlp{8cm}}
\toprule
& \textbf{Category} & \textbf{Model} & \textbf{Description} \\
\midrule
\multirow{7}{*}{\rotatebox{90}{\tiny \textsc{\textcolor{gray!70!black}{Supervised Tabular ML~~~~~~~~~~~~~~~~~~~~~~~~~~~~~~~}}}}
& \emph{Gradient Boosted} & XGBoost~\tiny\citep{xgboost} & \multirow{3}{*}{\parbox[t]{8cm}{Ensemble of decision trees trained via gradient boosting, where each tree corrects the residual errors of the previous ones, forming a competitive baseline on tabular data.}} \\
& \emph{Decision Tree} & LightGBM~\tiny\citep{lightgbm} \\
& & CatBoost~\tiny\citep{catboost} \\
& \emph{Deep Tabular ML} & \parbox[t]{3cm}{CARTE\\\tiny\citep{carte}} & Deep representation learning on tabular data, trained in a (self-)supervised fashion. \\
& \emph{AutoML} & \parbox[t]{3cm}{AutoGluon\\\tiny\citep{autogluon}} & Ensemble of diverse learners (\emph{e.g.}, Neural Networks, GBDTs) with automated hyperparameter optimization. Generally
provides best performance among tabular models. \\
& \parbox[t]{3cm}{\raggedleft \emph{Manual Feature Engineering}} & \parbox[t]{3cm}{DS+LightGBM\\[1.5pt]DS+AutoGluon\\[5pt]DS+TabPFN-2.5\\...\tiny\citep{relbench}} & Current gold-standard. A data scientist expert that engineers task-specific features (\emph{e.g.}, temporal aggregations, multi-hop joins) to capture relational dependencies, followed by a tabular model.  \\
& \parbox[t]{3cm}{\raggedleft \emph{Automatic Feature Engineering}} & \parbox[t]{3cm}{DFS+XGBoost\\[4pt]DFS+AutoGluon\\\tiny\citep{kanter2015deep}} & Task-agnostic, fixed-function, column-wise encoder to flatten multi-tables based on deep feature synthesis, followed by a tabular model. \\
\cdashline{2-4}
\multirow{6}{*}{\rotatebox{90}{\tiny \textsc{\textcolor{gray!70!black}{Sup. Relational ML~~}}}}
& \emph{Graph} & GraphSAGE~\tiny\citep{graphsage} & \multirow{4}{*}{\parbox[t]{8cm}{Message-passing networks based on different aggregation schemes and information flows, trained in a supervised fashion based on the relational deep learning paradigm.}} \\
& \emph{Neural Network} & GAT~\tiny\citep{gat} \\
& & PNA~\tiny\citep{pna} \\
& & RelGNN~\tiny\citep{relgnn} \\
& \emph{Graph Transformer} & HGT/HGT$_{\textrm{PE}}$~\tiny\citep{hgt} & \multirow{2}{*}{\parbox[t]{8cm}{Self-attention networks, trained in a supervised fashion based on the relational deep learning paradigm.}} \\
& & RelGT~\tiny\citep{relgt} \\
\cdashline{2-4}
\multirow{9}{*}{\rotatebox{90}{\tiny \textsc{\textcolor{gray!70!black}{Foundation Models~~~~~~~~~~~~~~~~~~~~~~~}}}}
& \emph{Language Model} & LLM$_1$~\tiny\citep{ranjan2026rt} & \multirow{2}{*}{\parbox[t]{8cm}{Language models adapted for relational reasoning through structured prompts and context encoding.}} \\
& & LLM$_2$~\tiny\citep{relllm} \\
& \emph{Tabular} & TabPFN-2.5~\tiny\citep{tabpfnv2} & \multirow{2}{*}{\parbox[t]{8cm}{In-context learning over single tabular data based on a pre-trained transformer architecture.}} \\
& \emph{Foundation Model} & TabICLv2~\tiny\citep{qu2026tabiclv2} \\
& \parbox[t]{3cm}{\raggedleft \emph{Relational Foundation Model}} & \parbox[t]{3cm}{Griffin\\\tiny\citep{griffin}} & Pre-trained relational foundation model that leverages large-scale meta-learning. \\
& & \parbox[t]{3cm}{RT$_{\textrm{zero}}$\\\tiny\citep{ranjan2026rt}} & Relational transformer for ``zero''-shot predictions, pre-trained via masked token prediction over relational data. \\
& & \parbox[t]{3cm}{GNN+TabPFN-2.5\\\tiny\citep{gnntabpfn}} & Randomly initialized GNN to flatten multi-tables, followed by a tabular foundation model. \\
& & \parbox[t]{3cm}{RDBLearn\\\tiny\citep{xu2026rdblearn}} & Task-agnostic deep feature synthesis, followed by a tabular foundation model. \\
& & \nameone~\tiny\citep{fey2025kumorfm} & The first iteration of \name. \\
\bottomrule
\end{tabular}
}
\end{table}

We compare \nametwo against a diverse set of baselines, including supervised tabular models, state-of-the-art relational deep learning models, as well as language-based, tabular, and relational foundation models.
To ensure strong and well-tuned baseline performance, we rely on publicly reported benchmark numbers for baselines as much as possible.
Baselines evaluated by ourselves are marked with $^*$.
Detailed information about all baselines is provided in Table~\ref{tab:baselines}.

Specifically, we evaluate tabular (foundation) models in three settings.
Unless stated otherwise, tabular models are run on entity table features only, without access to information from related tables.
This setting highlights the limitations of tabular models in capturing relational and temporal dependencies without additional feature engineering.
To address this, we consider two extended settings for applying tabular models to relational data.
First, we leverage an expert data scientist (DS) with MSc degree, 4.0 GPA and 5 years of experience of building machine learning models to manually design features (\emph{e.g.}, temporal aggregations, multi-hop joins) for specific tasks, incorporated into a single wide table~\citep{relbench}.
Since this process needs to be repeated individually for each task and requires hours to days of human labor, it is not available for all tasks.
Secondly, we explore automatic feature engineering pipelines via fixed-function encoders, \emph{e.g.}, based on the deep feature synthesis (DFS) framework~\citep{kanter2015deep} or via randomly initialized graph neural networks (GNNs)~\citep{gnntabpfn}.
Afterwards, these single wide tables can be fed into any tabular model, denoted by prepending DS+ and DFS+ to the tabular model, respectively. Notably, RDBLearn can be interpreted as DFS+TabPFN-2.5~\citep{xu2026rdblearn}.

On the foundation model side, we compare against language-based models, tabular foundation models (following the recipe above), and dedicated relational foundation models.
Language models~\citep{relllm,ranjan2026rt} are prompted with task instructions followed by a series of text-serialized database subgraphs and their targets used as context examples for in-context learning.
Relational foundation baselines, such as Griffin~\citep{griffin}, RT$_{\textrm{zero}}$~\citep{ranjan2026rt} or \nameone~\citep{fey2025kumorfm}, are specifically pre-trained end-to-end on relational data.

\begin{table}
\centering
\caption{\textbf{Test results on the binary classification tasks in RelBenchV1.} Higher is better (AUROC). \textcolor{green}{\raisebox{-0.2ex}{\rule{0.8em}{0.8em}}}~denotes the best, and \textcolor{green!50!white}{\raisebox{-0.2ex}{\rule{0.8em}{0.8em}}}~the second-best among all models and tasks.}
\label{tab:rbv1-cls}
\setlength{\tabcolsep}{2pt}
\resizebox{\textwidth}{!}{
\begin{tabular}{llrrrrrrrrrrrrrr}
\toprule
& \multirow{2}{*}{\textbf{Method}} & \multicolumn{2}{c}{\texttt{\small f1}} & \multicolumn{2}{c}{\texttt{\small avito}} & \multicolumn{2}{c}{\texttt{\small event}} & \multicolumn{1}{c}{\texttt{\small trial}} & \multicolumn{2}{c}{\texttt{\small amazon}} & \multicolumn{2}{c}{\texttt{\small stack}} & \multicolumn{1}{c}{\texttt{\small hm}} & \multicolumn{1}{c}{\textbf{\small Avg}} & \multicolumn{1}{c}{\textbf{\small Rank}} \\
& & \multicolumn{1}{c}{\texttt{\scriptsize dnf}} & \multicolumn{1}{c}{\texttt{\scriptsize top3}} & \multicolumn{1}{c}{\texttt{\scriptsize click}} & \multicolumn{1}{c}{\texttt{\scriptsize visit}} & \multicolumn{1}{c}{\texttt{\scriptsize repeat}} & \multicolumn{1}{c}{\texttt{\scriptsize ignore}} & \multicolumn{1}{c}{\texttt{\scriptsize out}} & \multicolumn{1}{c}{\texttt{\scriptsize user}} & \multicolumn{1}{c}{\texttt{\scriptsize item}} & \multicolumn{1}{c}{\texttt{\scriptsize eng}} & \multicolumn{1}{c}{\texttt{\scriptsize badge}} & \multicolumn{1}{c}{\texttt{\scriptsize churn}} & \multicolumn{1}{c}{\scriptsize $\uparrow$} & \multicolumn{1}{c}{\scriptsize $\downarrow$} \\
\midrule
\multirow{4}{*}{\rotatebox{90}{\tiny \textsc{\textcolor{gray!70!black}{Sup. TabML~~}}}}
& \href{https://arxiv.org/abs/2407.20060}{LightGBM} & 68.56 & 73.92 & 53.60 & 53.05 & 68.04 & 79.93 & 70.09 & 52.22 & 62.54 & 63.39 & 63.43 & 55.21 & 63.67 & 16.75 \\
& \href{https://arxiv.org/abs/2407.20060}{DS+LightGBM} & 69.80 & 82.40 & 64.27 & 64.46 & 75.42 & 84.23 & 72.00 & 67.60 & 81.80 & 90.30 & 86.20 & 69.00 & 75.62 & 7.75 \\
& \href{https://arxiv.org/abs/2003.06505}{DS+AutoGluon}$^*$ & 70.01 & 82.53 & 65.66 & 65.87 & 74.85 & 82.31 & 70.51 & 68.12 & 80.45 & 89.94 & 85.38 & 68.71 & 75.36 & 8.50 \\
& \href{https://arxiv.org/abs/2511.08667}{DS+TabPFN-2.5}$^*$ & 71.16 & 77.70 & 64.21 & 64.53 & 76.43 & 85.70 & 70.75 & 66.28 & 79.83 & 89.07 & 85.08 & 68.31 & 74.92 & 9.50 \\
\cdashline{2-16}
\multirow{5}{*}{\rotatebox{90}{\tiny \textsc{\textcolor{gray!70!black}{Sup. RelationalML~~}}}}
& \href{https://arxiv.org/abs/2407.20060}{GraphSAGE} & 72.62 & 75.54 & 65.90 & 66.20 &
76.89 & 81.62 & 68.60 & \cellsecond 70.42 & \cellfirst 82.81 & \cellsecond 90.59 & \cellsecond 88.86 & \cellsecond 69.88 & 75.83 & 6.42 \\
& \href{https://arxiv.org/abs/2003.01332}{HGT} & 70.77 & 70.75 & 63.76 & 64.32 & 64.96 & 82.47 & 58.37 & 66.43 & 77.97 & 88.47 & 86.08 & 66.95 & 71.78 & 12.50 \\
& \href{https://arxiv.org/abs/2003.01332}{HGT$_{\textrm{PE}}$} & 71.17 & 76.27 & 64.57 & 64.95 & 65.36 & 81.61 & 59.21 & 66.19 & 78.03 & 88.17 & 85.66 & 65.69 & 72.24 & 11.75 \\
& \href{https://arxiv.org/abs/2502.06784}{RelGNN} & 75.29 & 85.69 & 68.23 & 66.18 & \cellsecond 79.61 & 86.18 & 71.24 & \cellfirst 70.99 & \cellsecond 82.64 & \cellfirst 90.75 & \cellfirst 88.98 & \cellfirst 70.93 & \cellsecond \textbf{78.06} & \cellsecond \textbf{3.75} \\
& \href{https://arxiv.org/abs/2505.10960}{RelGT} & 75.87 & 83.52 & \cellsecond 68.30 & \cellsecond 66.78 & 76.09 & 81.57 & 68.61 & 70.39 & 82.55 & 90.53 & 86.32 & 69.27 & 76.65 & 6.00 \\
\cdashline{2-16}
\multirow{9}{*}{\rotatebox{90}{\tiny \textsc{\textcolor{gray!70!black}{foundational~~}}}}
& \href{https://arxiv.org/abs/2510.06377}{LLM$_1$} & 75.80 & \cellsecond 91.40 & 59.80 & 62.70 & 71.40 & 69.30 & 57.40 & 58.10 & 62.10 & 78.00 & 80.00 & 59.80 & 68.82 & 13.92 \\
& \href{https://arxiv.org/abs/2411.11829}{LLM$_2$} & 80.03 & 87.11 & 53.36 & 54.07 & 70.11 & 68.65 & 59.17 & 62.55 & 73.41 & 81.23 & 79.99 & 63.81 & 69.46 & 13.92 \\
& \href{https://arxiv.org/abs/2511.08667}{TabPFN-2.5}$^*$ & 61.21 & 77.35 & 54.44 & 51.45 & 69.10 & 70.91 & \cellsecond 71.74 & 57.14 & 59.30 & 59.66 & 52.26 & 54.99 & 61.63 & 16.42 \\
& \href{https://arxiv.org/abs/2505.05568}{Griffin}& 57.70 & 82.50 & 45.90 & 60.70 & 71.88 & 83.27 & 51.00 & 62.30 & 69.00 & 77.50 & 73.50 & 60.20 & 66.29 & 15.42 \\
& \href{https://arxiv.org/abs/2510.06377}{RT$_{\textrm{zero}}$} & 81.20 & 89.30 & 59.50 & 61.80 & 73.22 & 77.47 & 51.80 & 64.00 & 70.90 & 75.70 & 80.10 & 62.80 & 67.73 & 13.25 \\
& \href{https://www.linkedin.com/posts/tomaszpalczewski_aaai-26-summer-symposium-series-ugcPost-7435006760857415681-xTpy}{GNN+TabPFN} & 64.26 & 74.67 & 60.68 & 65.35 & 73.58 & 84.13 & 56.68 & 64.30 & 77.89 & 87.48 & 85.34 & 64.34 & 71.56 & 12.58 \\
& \href{https://arxiv.org/abs/2602.13697}{RDBLearn} & 70.87 & 79.69 & \cellfirst 69.04 & 65.49 & 75.04 & 82.52 & 71.58 & 67.57 & 82.07 & 89.39 & 85.26 & 68.05 & 75.97 & 8.08 \\
& \href{https://kumo.ai/research/kumo_relational_foundation_model.pdf}{\nameone{}} & \cellsecond 82.41 & 91.07 & 64.85 & 64.11 & 76.08 & \cellsecond 89.20 & 70.79 & 67.29 & 79.93 & 87.09 & 80.00 & 67.71 & 76.71 & 8.33 \\
& \textbf{\nametwo{}} & \cellfirst 84.59 & \cellfirst 92.18 & 67.42 & \cellfirst 69.41 & \cellfirst 81.66 & \cellfirst 90.83 & \cellfirst 72.03 & 69.10 & 82.17 & 89.40 & 87.15 & 69.27 & \cellfirst \textbf{79.60} & \cellfirst \textbf{3.08} \\
\bottomrule
\end{tabular}}
\end{table}

\subsection{\nametwo on Binary Classification Tasks}
\label{subsec:binary_classification}

We evaluate \nametwo on the binary classification tasks of RelBenchV1, RelBenchV2 and 4DBInfer, comparing against data scientist-engineered tabular models, recent relational deep learning models, and various paradigms of foundation models.
All models follow the pre-defined benchmark splits, but may operate on different kinds of input representations, including single tables, flattened feature-engineered tables, or relational subgraphs.
All relational deep learning models, including \nametwo, use two-hop subgraphs to ensure a fair comparison.

Table~\ref{tab:rbv1-cls} presents the results on the 12 RelBenchV1 binary classification tasks. \nametwo achieves the highest average AUROC of 79.60 and the best overall rank of 3.08 among all evaluated methods, outperforming the best tabular model by 3.98 points, the previously best foundation model \mbox{\nameone} by 2.89 points, and even the best supervised relational model RelGNN by 1.54 points.

\paragraph{Comparison to Tabular Models.}

As expected, tabular models such as LightGBM or TabPFN-2.5, without additional engineered features (\emph{e.g.}, ``time since last purchase''), fail to capture predictive signals in these inherently temporal tasks, consistently ranking lowest across all models.
With data scientist–engineered features, tabular models achieve substantially stronger performance and become competitive with supervised relational models. In this setting, DS+LightGBM, DS+AutoGluon, and DS+TabPFN-2.5 perform similarly on average, with minor variations across tasks.
However, no such model consistently outperforms relational approaches, suggesting that they do not capture additional predictive signals beyond what is already learned by end-to-end relational models.
Moreover, overall performance does not reflect the significant manual effort required for feature engineering.
In contrast, \nametwo captures complex interactions and temporal patterns without manual labor, improving results by 3.98 points. This ability to maintain strong performance across diverse tasks without task-specific engineering and tuning highlights the effectiveness of our relational foundation approach.

\paragraph{Comparison to Supervised Relational Models.}

\nametwo outperforms even the strongest supervised baseline by 1.54 points on average across all tasks.
To our knowledge, this is the first demonstration of a few-shot foundation model surpassing supervised approaches on relational benchmarks.
Notably, this is achieved without any task-specific training and with as little as 0.2\% of the available data (context examples \emph{vs.} full training set), making it significantly more practical than supervised methods.
\nametwo is particularly effective in smaller training data regimes (\texttt{f1}, \texttt{avito}, \texttt{event}, \texttt{trial}), where it consistently outperforms any supervised approach, often by large margins.
We observe that supervised relational models retain a slight advantage on \texttt{amazon}, \texttt{stack}, and \texttt{hm} due to their larger training set sizes, where task-specific tuning can still provide incremental gains.
We expect this gap to diminish or reverse when scaling context sizes (\emph{cf.}~Sec.~\ref{subsec:ablations}).
Notably, while supervised relational models eliminated the need for manual feature engineering, \nametwo now eliminates the need for additional training, which can range from hours to days.

\paragraph{Comparison to LLMs.}

\nametwo consistently outperforms LLM-based baselines on predictive tasks.
Overall, LLMs underperform in this setting, as they are not designed for structured prediction and lack appropriate inductive biases for tabular and relational data.
We observe competitive performance only in cases with potential data leakage (\emph{e.g.}, \texttt{f1}) due to memorization effects in large pre-trained models.
Even in these scenarios, \nametwo significantly outperforms LLM-based approaches, demonstrating the advantage of models explicitly designed for relational prediction.

\paragraph{Comparison to Relational Foundation Models.}

Existing relational foundation models such as Griffin~\citep{griffin} and RT$_{\textrm{zero}}$~\citep{ranjan2026rt} generally tend to underperform on RelBenchV1 despite their strong relational prior.
A key limitation lies in their formulation of in-context learning as \emph{label propagation}, which restricts effective context utilization and often reduces predictions to its historical mean.
Fixed-function, training-free encoder pipelines as used in GNN+TabPFN~\citep{gnntabpfn} and RDBLearn~\citep{xu2026rdblearn} instead rely on automatically flattening relational data into single wide tables, effectively deferring most of the modeling capacity to the underlying tabular foundation model.
As a result, their performance remains bounded by that of feature-engineered tabular approaches, lacking the ability to perform explicit relational reasoning (\emph{cf.}~Sec.~\ref{sec:kumorfmv2}).
In contrast, \nametwo directly models relational structure and captures complex dependencies beyond column-level aggregations, resulting in average improvements of 8.04 points and 3.63 points in AUROC over these approaches.
Remarkably, \nametwo represents a significant improvement over \nameone.
On average, \nametwo outperforms v\onezero by 2.89 points, with consistent gains over all tasks.

\begin{wraptable}{r}{0.66\textwidth}
\centering
\vspace{-0.69cm}
\caption{\textbf{Test results on the binary classification tasks in RelBenchV2.} Higher is better (AUROC). \textcolor{green}{\raisebox{-0.2ex}{\rule{0.8em}{0.8em}}}~denotes the best, and \textcolor{green!50!white}{\raisebox{-0.2ex}{\rule{0.8em}{0.8em}}}~the second-best among all models and tasks.}
\label{tab:rbv2-cls}
\setlength{\tabcolsep}{3pt}
\resizebox{0.66\textwidth}{!}{
\begin{tabular}{lrrrrrrr}
\toprule
\multirow{2}{*}{\textbf{Method}} & \multicolumn{1}{c}{\texttt{\small mimic}} & \multicolumn{3}{c}{\texttt{\small ratebeer}} & \multicolumn{1}{c}{\texttt{\small arxiv}} & \multicolumn{1}{c}{\textbf{\small Avg}} & \multicolumn{1}{c}{\textbf{\small Rank}} \\
& \multicolumn{1}{c}{\texttt{\scriptsize stay}} & \multicolumn{1}{c}{\texttt{\scriptsize beer-churn}} & \multicolumn{1}{c}{\texttt{\scriptsize user-churn}} & \multicolumn{1}{c}{\texttt{\scriptsize dormant}} &
\multicolumn{1}{c}{\texttt{\scriptsize citation}} &
\multicolumn{1}{c}{\scriptsize $\uparrow$} & \multicolumn{1}{c}{\scriptsize $\downarrow$} \\
\midrule
\href{https://arxiv.org/abs/2602.12606}{LightGBM} & 51.81 & 76.21 & 83.92 & 75.79 & 71.21 & 71.79 & 3.80 \\
\href{https://arxiv.org/abs/2602.12606}{GraphSAGE} & 55.01 & \cellsecond 78.67 & \cellsecond 94.27 & \cellsecond 80.51 & \cellfirst 82.50 & \cellsecond \textbf{78.19} & \cellsecond \textbf{2.00} \\
\href{https://kumo.ai/research/kumo_relational_foundation_model.pdf}{\nameone{}} & \cellfirst 56.33 & 75.06 & 91.38 & 77.10 & 80.62 & 76.22 & 2.80 \\
\textbf{\nametwo{}} & \cellsecond 55.28 & \cellfirst 83.84 & \cellfirst 97.43 & \cellfirst 80.65 & \cellsecond 81.71 & \cellfirst \textbf{79.96} & \cellfirst \textbf{1.75} \\
\bottomrule
\end{tabular}
}
\end{wraptable}

\paragraph{\nametwo on RelBenchV2.}

We further evaluate \nametwo on RelBenchV2 (\emph{cf.}~Table~\ref{tab:rbv2-cls}).
Due to the recency of the benchmark, publicly reported results are limited.
We compare against a subset of supervised tabular and relational approaches, as well as to \nameone.
Again, \nametwo achieves the highest overall performance, surpassing LightGBM by 8.17 points, GraphSAGE by 1.77 points, and \nameone by 3.74 points.
Notably, \nametwo performs strongest on all tasks of the \texttt{ratebeer} database, despite it having the largest training set.
We also observe substantial gains over \nameone on this dataset.
As \texttt{ratebeer} contains the highest number of columns (221), this highlights the importance of task-conditioned feature processing to effectively leverage high-dimensional inputs.

\paragraph{\nametwo on 4DBInfer.}

\begin{wraptable}{r}{0.62\textwidth}
\centering
\vspace{-0.68cm}
\caption{\textbf{Test results on the binary classification tasks in 4DBInfer.} Higher is better (AUROC). \textcolor{green}{\raisebox{-0.2ex}{\rule{0.8em}{0.8em}}}~denotes the best, and \textcolor{green!50!white}{\raisebox{-0.2ex}{\rule{0.8em}{0.8em}}}~the second-best among all models and tasks.}
\label{tab:dbinfer}
\setlength{\tabcolsep}{3pt}
\resizebox{0.62\textwidth}{!}{
\begin{tabular}{llrrrrrrr}
\toprule
& \multirow{2}{*}{\textbf{Method}} & \multicolumn{1}{c}{\small AB} & \multicolumn{1}{c}{\small OB} & \multicolumn{1}{c}{\small RR} & \multicolumn{2}{c}{\small SE} & \multicolumn{1}{c}{\textbf{\small Avg}} & \multicolumn{1}{c}{\textbf{\small Rank}} \\
& & \multicolumn{1}{c}{\texttt{\scriptsize churn}} & \multicolumn{1}{c}{\texttt{\scriptsize ctr}} & \multicolumn{1}{c}{\texttt{\scriptsize cvr}} & \multicolumn{1}{c}{\texttt{\scriptsize upvote}} & \multicolumn{1}{c}{\texttt{\scriptsize churn}} & \multicolumn{1}{c}{\scriptsize $\uparrow$} & \multicolumn{1}{c}{\scriptsize $\downarrow$} \\
\midrule
\multirow{4}{*}{\rotatebox{90}{\tiny \textsc{\textcolor{gray!70!black}{Sup. TabML~~}}}}
& \href{https://arxiv.org/abs/2404.18209}{XGBoost} & 50.00 & 50.00 & 50.00 & 49.68 & 50.84 & 50.10 & 10.6 \\
& \href{https://arxiv.org/abs/2404.18209}{AutoGluon} & 50.00 & 49.69 & 50.96 & 50.81 & 50.00 & 50.29 & 10.4 \\
& \href{https://arxiv.org/abs/2404.18209}{DFS+XGBoost} & 69.22 & 54.21 & 79.06 & 86.75 & 82.51 & 74.35 & 9.0 \\
& \href{https://arxiv.org/abs/2404.18209}{DFS+AutoGluon} & 72.91 & 54.94 & 80.08 & 88.49 & 83.96 & 76.08 & 7.6 \\
\cdashline{2-9}
\multirow{4}{*}{\rotatebox{90}{\tiny \textsc{\textcolor{gray!70!black}{Sup. RelML~~}}}}
& \href{https://arxiv.org/abs/2404.18209}{GAT} & 76.22 & 61.46 & 82.84 & 88.53 & 86.45 & 79.10 & 5.8\\
& \href{https://arxiv.org/abs/2404.18209}{HGT} &\cellsecond 77.30 & \cellfirst 62.60 & \cellfirst 84.95 & 88.17 & 86.70 & \cellsecond 79.94 &\cellfirst 3.4 \\
& \href{https://arxiv.org/abs/2404.18209}{PNA} & 76.45 & \cellsecond 62.49 & 83.67 &\cellfirst 88.96 & 86.64 & 79.64 & 4.0 \\
& \href{https://arxiv.org/abs/2404.18209}{GraphSAGE} & 75.71 & 62.39 & \cellsecond 84.70 & 88.61 & 85.58 & 79.40 & 4.6 \\
\cdashline{2-9}
\multirow{3}{*}{\rotatebox{90}{\tiny \textsc{\textcolor{gray!70!black}{FDN~~}}}}
& \href{https://arxiv.org/abs/2602.13697}{RDBLearn} & \cellfirst 77.41 & 54.47 & 84.69 & 88.45 &\cellfirst 87.96 & 78.60 & 3.8 \\
& \href{https://kumo.ai/research/kumo_relational_foundation_model.pdf}{\nameone{}} & 75.59 & 61.11 & 84.36 &\cellsecond 88.92 & 87.87 & 79.17 & 4.6 \\
& \textbf{\nametwo{}} & 77.06 & 61.97 & 84.36 & 88.50 &\cellsecond 87.92 & \cellfirst 79.96 &\cellsecond 3.6 \\
\bottomrule
\end{tabular}}
\end{wraptable}

For the 4DBInfer benchmark, we report AUROC results across five binary classification tasks, with supervised baselines sourced from \citet{wang2024dbinfer} (\emph{cf.}~Table~\ref{tab:dbinfer}).
We compare against tabular models with and without automatic feature engineering (DFS+XGBoost and DFS+AutoGluon), as well as supervised relational models and few-shot foundation models.

\nametwo performs effectively on par with the best supervised relational models, achieving the highest average performance (79.96 AUROC) and a top overall rank.
While it does not dominate any single task, it consistently ranks among the top performers, indicating strong robustness across diverse datasets.
In contrast, the best-performing model varies across tasks, suggesting that no single approach generalizes uniformly well.

Notably, \nametwo matches or exceeds competitive relational baselines on most tasks while avoiding task-specific training. This consistency is particularly evident when compared to models such as RDBLearn or \nameone, which achieve strong performance on individual tasks but exhibit larger variability overall.
Furthermore, approaches based on deep feature synthesis show weaker performance on this benchmark, suggesting the benefits of end-to-end relational modeling.

\subsection{\nametwo on Multi-Class Classification Tasks}
\label{subsec:multi_classification}

\begin{table}[b]
\centering
\caption{\textbf{Test results on the multi-class classification tasks in SALT.} Higher is better (MRR). \textcolor{green}{\raisebox{-0.2ex}{\rule{0.8em}{0.8em}}}~denotes the best, and \textcolor{green!50!white}{\raisebox{-0.2ex}{\rule{0.8em}{0.8em}}}~the second-best among all models and tasks.}
\label{tab:salt}
\setlength{\tabcolsep}{3pt}
\resizebox{0.9\textwidth}{!}{
\begin{tabular}{llrrrrrrrrrr}
\toprule
& \multirow{2}{*}{\textbf{Method}} & \multicolumn{1}{c}{\multirow{2}{*}{\small Plant}} & \multicolumn{1}{c}{\small Shipping} & \multicolumn{1}{c}{\small Item} & \multicolumn{1}{c}{\small Header} & \multicolumn{1}{c}{\small Sales} & \multicolumn{1}{c}{\small Sales} & \multicolumn{1}{c}{\small Payment} & \multicolumn{1}{c}{\small Shipping} & \multicolumn{1}{c}{\textbf{\small Avg}} & \multicolumn{1}{c}{\textbf{\small Rank}} \\
& & & \multicolumn{1}{c}{\small Point}& \multicolumn{1}{c}{\small Incoterm} & \multicolumn{1}{c}{\small Incoterm} & \multicolumn{1}{c}{\small Office} & \multicolumn{1}{c}{\small Group} & \multicolumn{1}{c}{\small Terms} & \multicolumn{1}{c}{\small Condition} & \multicolumn{1}{c}{\scriptsize $\uparrow$} & \multicolumn{1}{c}{\scriptsize $\downarrow$} \\
\midrule
\multirow{2}{*}{\rotatebox{90}{\tiny \textsc{\textcolor{gray!70!black}{Base~~}}}}
& \href{https://arxiv.org/abs/2501.03413}{Random} & 0.59 & 0.11 & 0.62 & 0.62 & 0.99 & 0.02 & 0.02 & 0.12 & 0.39 & 10.38 \\
& \href{https://arxiv.org/abs/2501.03413}{Majority} & 0.59 & 0.54 & 0.62 & 0.62 & 0.99 & 0.05 & 0.23 & 0.41 & 0.51 & 9.63 \\
\cdashline{2-12}
& \href{https://arxiv.org/abs/2501.03413}{DS+XGBoost} & \cellfirst 0.99 & 0.95 & 0.70 & 0.70 & 0.99 & \cellsecond 0.51 & 0.57 & 0.68 & 0.76 & 5.25 \\
\multirow{3}{*}{\rotatebox{90}{\tiny \textsc{\textcolor{gray!70!black}{Sup. TabML~~}}}}
& \href{https://arxiv.org/abs/2501.03413}{DS+LightGBM} & 0.61 & 0.28 & 0.73 & 0.73 & 0.99 & 0.02 & 0.10 & 0.51 & 0.50 & 9.00 \\
& \href{https://arxiv.org/abs/2501.03413}{DS+CatBoost} & \cellfirst 0.99 & \cellfirst 0.80 & \cellfirst 0.80 & \cellfirst 0.81 & 0.99 & 0.16 & 0.44 & 0.71 & 0.71 & 4.63 \\
& \href{https://arxiv.org/abs/2501.03413}{DS+CARTE} & \cellfirst 0.99 & 0.97 & 0.75 & 0.77 & 0.99 & 0.46 & 0.62 & 0.74 & 0.79 & 3.88 \\
\multirow{4}{*}{\rotatebox{90}{\tiny \textsc{\textcolor{gray!70!black}{Sup. RelML~~}}}}
& \href{https://arxiv.org/abs/2501.03413}{DS+AutoGluon} & \cellfirst 0.99 & \cellsecond 0.98 & 0.78 & 0.78 & 0.99 & 0.34 & 0.52 & 0.74 & 0.77 & 3.75 \\
\cdashline{2-12}
& \href{https://arxiv.org/abs/2501.03413}{GraphSAGE$_1$} & \cellfirst 0.99 & 0.97 & 0.64 & 0.59 & 0.99 & 0.20 & 0.39 & 0.59 & 0.67 & 6.88 \\
& \href{https://arxiv.org/abs/2602.12606}{GraphSAGE$_2$} & \cellfirst 0.99 & \cellsecond 0.98 & 0.69 & 0.62 & \cellfirst 1.00 & 0.16 & 0.37 & 0.57 & 0.67 & 5.88 \\
\cdashline{2-12}
& \href{https://arxiv.org/abs/2602.11139}{DS+TabICLv2}$^*$ & 0.94 & 0.91 & \cellfirst 0.80 & 0.74 & \cellfirst 1.00 & 0.07 & 0.50 & 0.73 & 0.71 & 5.63 \\
\multirow{1}{*}{\rotatebox{90}{\tiny \textsc{\textcolor{gray!70!black}{FDN~}}}}
& \href{https://kumo.ai/research/kumo_relational_foundation_model.pdf}{\nameone{}} & \cellfirst 0.99 & \cellfirst 0.99 & 0.79 & \cellfirst 0.81 & \cellfirst 1.00 & 0.38 & \cellsecond 0.66 & \cellsecond 0.78 & \cellsecond \textbf{0.80} & \cellsecond \textbf{1.88} \\
& \textbf{\nametwo{}} & \cellfirst 0.99 & \cellsecond 0.98 & 0.78 & \cellfirst 0.81 & \cellfirst 1.00 & \cellfirst 0.61 & \cellfirst 0.68 & \cellfirst 0.79 & \cellfirst \textbf{0.83} & \cellfirst \textbf{1.38} \\
\bottomrule
\end{tabular}}
\end{table}

We now evaluate \nametwo in the multi-class classification setting on the SALT benchmark suite.
This enterprise resource planning dataset comprises 8 predictive tasks, characterized by significant class imbalances, a wide range of class counts (up to 589 for the ``Sales Group'' task), and challenges such as label diversity, noise, and distributional drift.
Notably, SALT is provided in both multi-table and joined single-table formats, making it an ideal testbed for both tabular and relational models.
One key challenge for foundation models specifically is the large class count, in which close to 600 classes need to be inferred based of only $10k$ context examples.
To address this, we adopt a hierarchical classification strategy~\citep{hierachicalclassification} similar to recent tabular foundation models.

Results are reported in Table~\ref{tab:salt} using Mean Reciprocal Rank (MRR).
\nametwo{} performs strongly out-of-the-box, outperforming giant tabular model ensembles such as AutoGluon by 6 points, as well as pre-trained and fine-tuned tabular models such as CARTE by 4 points.
Notably, \nametwo{} outperforms its closest competitor CARTE on all tasks.
We observe particularly strong gains from \nameone to \nametwo on the ``Sales Group’’ task, with the largest number of classes. While v\onezero struggled to achieve competitive performance in this setting, \nametwo handles large multi-class problems much more effectively, massively outperforming both supervised and foundation models.

\subsection{\nametwo on Regression Tasks}
\label{subsec:regression}

\begin{table}[t]
\centering
\caption{\textbf{Test results on the regression tasks in RelBenchV1.} Lower is better (MAE). Average performance $\bm{\mu}_n$ is normalized relative to the LightGBM baseline. \textcolor{green}{\raisebox{-0.2ex}{\rule{0.8em}{0.8em}}}~denotes the best, and \textcolor{green!50!white}{\raisebox{-0.2ex}{\rule{0.8em}{0.8em}}}~the second-best among all models and tasks.}
\label{tab:rbv1-reg}
\setlength{\tabcolsep}{3pt}
\resizebox{0.95\textwidth}{!}{
\begin{tabular}{llrrrrrrrrrrr}
\toprule
& \multirow{2}{*}{\textbf{Method}} & \multicolumn{1}{c}{\texttt{\small f1}} & \multicolumn{1}{c}{\texttt{\small avito}} & \multicolumn{1}{c}{\texttt{\small event}} & \multicolumn{2}{c}{\texttt{\small trial}} & \multicolumn{2}{c}{\texttt{\small amazon}} & \multicolumn{1}{c}{\texttt{\small stack}} & \multicolumn{1}{c}{\texttt{\small hm}} & \multicolumn{1}{c}{\bm{$\mu$}$_n$} & \multicolumn{1}{c}{\textbf{\small Rank}} \\
& & \multicolumn{1}{c}{\texttt{\scriptsize pos}} & \multicolumn{1}{c}{\texttt{\scriptsize ctr}} & \multicolumn{1}{c}{\texttt{\scriptsize attend}} & \multicolumn{1}{c}{\texttt{\scriptsize adverse}} & \multicolumn{1}{c}{\texttt{\scriptsize success}} & \multicolumn{1}{c}{\texttt{\scriptsize user}} & \multicolumn{1}{c}{\texttt{\scriptsize item}} & \multicolumn{1}{c}{\texttt{\scriptsize votes}} & \multicolumn{1}{c}{\texttt{\scriptsize sales}} & \multicolumn{1}{c}{\scriptsize $\downarrow$} & \multicolumn{1}{c}{\scriptsize $\downarrow$} \\
\midrule
\multirow{2}{*}{\rotatebox{90}{\tiny \textsc{\textcolor{gray!70!black}{Base~~}}}}
& \href{https://arxiv.org/abs/2407.20060}{Global Median} & 4.399 & 0.043 & 0.264 & 57.533 & 0.462 & 16.783 & 64.234 & 0.068 & 0.076 & 1.062 & 12.78 \\
& \href{https://arxiv.org/abs/2407.20060}{Entity Median} & 8.519 & 0.046 & 0.269 & 57.930 & 0.441 & 17.423 & 66.436 & 0.069 & 0.078 & 1.190 & 14.00 \\
\cdashline{2-13}
\multirow{4}{*}{\rotatebox{90}{\tiny \textsc{\textcolor{gray!70!black}{Sup. TabML~~}}}}
& \href{https://arxiv.org/abs/2407.20060}{LightGBM} & 4.170 & 0.041 & 0.264 & 44.011 & 0.425 & 16.783 & 60.569 & 0.068 & 0.076 & 1.000 & 10.22 \\
& \href{https://arxiv.org/abs/2407.20060}{DS+LightGBM} & 3.963 & 0.044 & 0.284 & \cellfirst 40.581 & 0.407 & \cellsecond 13.928 & \cellfirst 41.122 & \cellsecond 0.065 & \cellsecond 0.036 & 0.879 & 5.11 \\
& \href{https://arxiv.org/abs/2003.06505}{DS+AutoGluon}$^*$ & 4.251 & 0.045 & 0.256 & 44.706 & 0.430 & 14.399 & \cellsecond 45.390 & 0.068 & 0.043 & 0.918 & 7.11 \\
& \href{https://arxiv.org/abs/2511.08667}{DS+TabPFN-2.5}$^*$ & 4.373 & 0.044 & 0.318 & 47.168 & 0.432 & 15.631 & 47.908 & 0.080 & 0.059 & 1.009 & 10.11 \\
\cdashline{2-13}
\multirow{5}{*}{\rotatebox{90}{\tiny \textsc{\textcolor{gray!70!black}{Sup. RelationalML~~}}}}
& \href{https://arxiv.org/abs/2407.20060}{GraphSAGE} & 4.022 & 0.041 & 0.258 & 44.473 & 0.400 & 14.313 & 50.053 & \cellsecond 0.065 & 0.056 & 0.918 & 6.33 \\
& \href{https://arxiv.org/abs/2003.01332}{HGT} & 4.226 & 0.046 & 0.264 & 45.169 & 0.443 & 15.412 & 55.868 & 0.068 & 0.064 & 0.988 & 10.44 \\
& \href{https://arxiv.org/abs/2003.01332}{HGT$_{\textrm{PE}}$} & 4.392 & 0.048 & 0.261 & \cellsecond 42.648 & 0.440 & 15.864 & 55.849 & 0.068 & 0.064 & 0.992 & 10.00 \\
& \href{https://arxiv.org/abs/2502.06784}{RelGNN} & 3.798 & 0.037 & 0.238 & 44.461 & \cellfirst 0.301 & 14.230 & 48.767 & \cellsecond 0.065 & 0.054 & \cellsecond \textbf{0.861} & \cellsecond \textbf{4.44} \\
& \href{https://arxiv.org/abs/2505.10960}{RelGT} & 3.917 & 0.035 & 0.250 & 43.992 & \cellsecond 0.326 & 14.267 & 48.922 & \cellsecond 0.065 & 0.054 & 0.869 & 4.67 \\
\cdashline{2-13}
\multirow{6}{*}{\rotatebox{90}{\tiny \textsc{\textcolor{gray!70!black}{foundational~~}}}}
& \href{https://arxiv.org/abs/2511.08667}{TabPFN-2.5}$^*$ & 4.446 & 0.044 & 0.527 & 55.187 & 0.469 & 17.513 & 60.996 & 0.104 & 0.096 & 1.258 & 14.78 \\
& \href{https://arxiv.org/abs/2505.05568}{Griffin} & 4.460 & 0.050 & 0.461 & 78.232 & 0.463 & 35.590 & 53.214 & 0.092 & 0.151 & 1.471 & 15.44 \\
& \href{https://arxiv.org/abs/2510.06377}{RT$_{\textrm{zero}}$} & 2.901  & 0.058  & 0.379 & 73.999 & 0.455 & 18.802 & 57.996 & 0.110  & 0.089 & 1.240 & 14.00 \\
& \href{https://arxiv.org/abs/2602.13697}{RDBLearn} & 3.834 & \cellfirst 0.034 & \cellsecond 0.237 & 43.913 & 0.424 & 14.540 & 48.559 & 0.068 & 0.064 & 0.906 & 5.89 \\
& \href{https://kumo.ai/research/kumo_relational_foundation_model.pdf}{\nameone{}} & \cellfirst 2.747 & 0.035 & 0.264 & 58.231 & 0.417 & 16.161 & 55.254 & \cellsecond 0.065 & 0.040 & 0.908 & 7.22 \\
& \textbf{\nametwo{}}& \cellsecond 2.854 & \cellfirst 0.034 & \cellfirst 0.235 & 43.293 & 0.433 & \cellfirst 13.921 & 46.992 & \cellfirst 0.064 & \cellfirst 0.034 & \cellfirst \textbf{0.822} & \cellfirst \textbf{2.67} \\
\bottomrule
\end{tabular}}
\end{table}

\begin{wraptable}{r}{0.505\textwidth}
\centering
\vspace{-0.68cm}
\caption{\textbf{Test results on the regression tasks in RelBenchV2.} Lower is better (MAE). Average performance $\bm{\mu}_n$ is normalized relative to the LightGBM baseline. \textcolor{green}{\raisebox{-0.2ex}{\rule{0.8em}{0.8em}}}~denotes the best, and \textcolor{green!50!white}{\raisebox{-0.2ex}{\rule{0.8em}{0.8em}}}~the second-best among all models and tasks.}
\label{tab:rbv2-reg}
\setlength{\tabcolsep}{3pt}
\resizebox{0.48\textwidth}{!}{
\begin{tabular}{lrrrr}
\toprule
\multirow{2}{*}{\textbf{Method}} & \multicolumn{1}{c}{\texttt{\small ratebeer}} & \multicolumn{1}{c}{\texttt{\small arxiv}} & \multicolumn{1}{c}{\bm{$\mu$}$_n$} & \multicolumn{1}{c}{\textbf{\small Rank}} \\
& \multicolumn{1}{c}{\texttt{\scriptsize user-count}} & \multicolumn{1}{c}{\texttt{\scriptsize publication}} & \multicolumn{1}{c}{\scriptsize $\downarrow$} & \multicolumn{1}{c}{\scriptsize $\downarrow$} \\
\midrule
\href{https://arxiv.org/abs/2602.12606}{Global Median} & 15.124 & 0.577 & 0.872 & 4.5 \\
\href{https://arxiv.org/abs/2602.12606}{Entity Median} & 13.079 & 0.874 & 1.079 & 5.0 \\
\href{https://arxiv.org/abs/2602.12606}{LightGBM} & 20.350 & 0.577 & 1.000 & 5.0 \\
\href{https://arxiv.org/abs/2602.12606}{GraphSAGE} & \cellsecond 7.374 & \cellsecond 0.513 & \cellsecond \textbf{0.626} & \cellsecond \textbf{2.0} \\
\href{https://kumo.ai/research/kumo_relational_foundation_model.pdf}{\nameone{}} & 11.063 & 0.518 & 0.712 & 3.0 \\
\textbf{\nametwo{}} & \cellfirst 7.298 & \cellfirst 0.487 & \cellfirst \textbf{0.601} & \cellfirst \textbf{1.0} \\
\bottomrule
\end{tabular}
}
\end{wraptable}

Finally, we evaluate \nametwo on the 11 regression tasks from RelBenchV1 and RelBenchV2 (\emph{cf.}~Tables~\ref{tab:rbv1-reg} and \ref{tab:rbv2-reg}).
We compare against a similar set of baselines and adopt the same experimental setup as in the binary classification experiments on RelBench (\emph{cf.}~Sec.~\ref{subsec:binary_classification}).
We report the additional ``Global Median'' and ``Entity Median'' baselines as they provide strong sanity checks regarding model performance.
For all experiments, we report task-specific Mean Absolute Error (MAE), and compute normalized average error $\mu_n$ relative to the naive LightGBM baseline (lower is better).

On RelBenchV1, \nametwo achieves the lowest normalized MAE ($\mu_n = 0.822$) and the best overall rank across the suite, outperforming the strongest baseline by 4.7\% on average. Furthermore, it obtains the top performance in 5 out of the 9 regression tasks.
Compared to \nameone, \nametwo improves performance by 10.5\%, representing a substantial gain.
On RelBenchV2, \nametwo again achieves the lowest score ($\mu_n = 0.601$) with a perfect rank of 1.0, while its predecessor \nameone trails behind the supervised GraphSAGE model, highlighting the improved competitiveness of \nametwo against strong supervised relational baselines.

Overall, regression tasks exhibit trends similar to those observed in classification on RelBench (\emph{cf.}~Sec.~\ref{subsec:binary_classification}).
One notable difference is the increased competitiveness of the data scientist-engineered pipelines.
These specific tasks are characterized by high absolute error, suggesting that in regimes where the target is highly volatile, expert-driven manual feature engineering can occasionally outperform automated graph-based approaches.
We attribute this to the difficulty of capturing long-term, coarse-grained statistical patterns from fine-grained relational subgraphs at the level of individual transactions.
\nametwo mitigates this limitation through its support for lagged targets within its smarter context selection strategy (\emph{cf.}~Sec.~\ref{sec:kumorfmv2}).

\subsection{Robustness of \nametwo}
\label{subsec:ablations}

\begin{figure}[t]
  \centering
  \raggedright
  \subfigure[Feature scarcity]{
    \includegraphics[height=3.4cm]{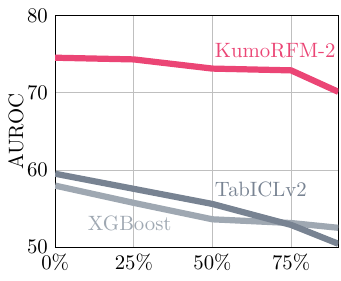}\label{fig:featdrop}
  }
  \hspace{0.15cm}
  \subfigure[Link scarcity]{
    \includegraphics[height=3.4cm]{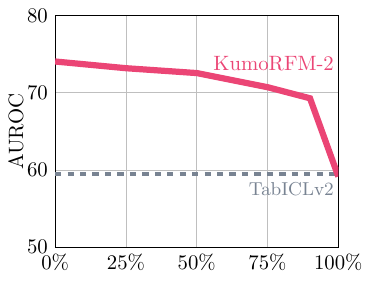}\label{fig:edgedrop}
  }
  \hspace{-0.32cm}
  \subfigure[Number of noisy columns]{
    \includegraphics[height=3.4cm]{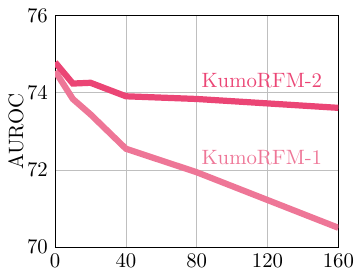}\label{fig:noise}
  }
  \caption{\textbf{Robustness of \nametwo to sparse, incomplete and noisy data}. \nametwo maintains strong resilience to data scarcity, and remains robust to noise compared to its predecessor.}\label{fig:ablation2}
\end{figure}

To better understand the performance and robustness of \nametwo, we conduct two sets of studies.
First, we analyze the key factors driving model performance, focusing on context size as well as the depth and breadth of relational subgraphs (\emph{cf}.~Fig.~\ref{fig:ablation1}).
Second, we evaluate model robustness under challenging conditions, including data scarcity (simulating cold-start scenarios) and increasing feature noise (\emph{cf.}~Fig.~\ref{fig:ablation2}).
The results demonstrate the model's ability to effectively leverage relational structure while remaining robust across diverse settings.
We now discuss each study in detail:

\begin{figure}[t]
  \centering
  \raggedright
  \subfigure[Context size]{
    \includegraphics[height=3.4cm]{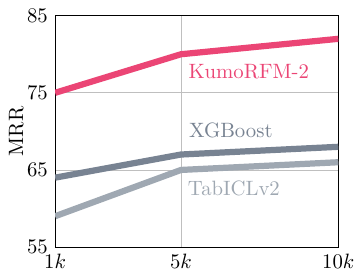}\label{fig:context}
  }
  \hspace{-0.17cm}
  \subfigure[Subgraph depth]{
    \includegraphics[height=3.4cm]{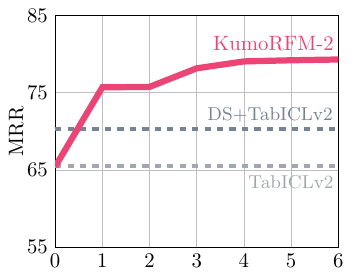}\label{fig:hop}
  }
    \hspace{-0.13cm}
  \subfigure[Subgraph breadth]{
    \includegraphics[height=3.4cm]{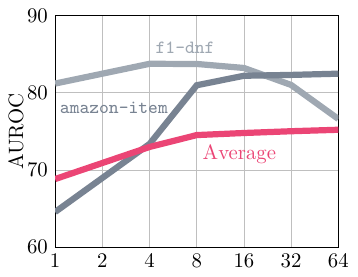}\label{fig:neighs-scale}
  }
  \caption{\textbf{Robustness of \nametwo with respect to context construction}. \nametwo exhibits strong sample efficiency, and generally benefits from increased relational subgraph breadth and depth.
  We identify subgraph breadth as a key factor for optimizing task-specific performance.
  }\label{fig:ablation1}
\end{figure}

\paragraph{\nametwo achieves strong sample efficiency and scales effectively.}

We evaluate average model performance on a subset of SALT under varying data regimes by adjusting the context size, \emph{i.e.} the number of training/context examples (\emph{cf.}~Fig.~\ref{fig:context}).
We compare against XGBoost (supervised) and TabICLv2 (in-context) baselines.
Overall, \nametwo exhibits strong sample efficiency, performing competitively even with small context sizes of $1k$ (0.05\% of training examples).
Furthermore, \nametwo follows similar scaling trends as XGBoost and TabICLv2, showing more rapid initial gains followed by more gradual improvements as context/supervision increases.
Notably, \nametwo continues to improve more consistently than the baselines at larger context sizes.
We attribute this to the richer context representations, where subgraphs capture more diverse relational structure than flat representations, and thus enable more effective use of larger contexts.

\paragraph{\nametwo benefits from increasing relational context.}

We evaluate average model performance on a subset of SALT under varying subgraph depths from zero hops (effectively reducing to a single table setting) up to six hops (\emph{cf}.~Fig.~\ref{fig:hop}).
We compare against TabICLv2 and DS+TabICLv2.
Notably, \nametwo matches the performance of TabICLv2 at zero hops, demonstrating strong performance as a single tabular foundation model even without relational information.
We observe a sharp performance increase when moving from zero to one hop, highlighting the importance of incorporating immediate relational context.
From three hops onward, subgraphs cover all available tables in the \textsc{SALT} dataset.
Importantly, performance continues to improve as deeper relationships are incorporated, eventually saturating at around four hops.

\paragraph{\nametwo can benefit from task-dependent neighborhood sizes.}

We evaluate model performance on a subset of RelBenchV1-classification under varying neighborhood sizes per hop, ranging from 1 to 64, while fixing the subgraph depth to two hops (\emph{cf.}~Fig.~\ref{fig:neighs-scale}).
We report both average performance across tasks and representative task-specific examples.
Overall, increasing the neighborhood size leads to steady improvements in average performance, indicating that broader subgraphs enable the model to aggregate more informative relational signals.
However, the optimal neighborhood size is inherently task-dependent.
For instance, performance on the \texttt{driver-dnf} task in \texttt{f1} peaks early with a highly localized neighborhood ($n=4$) and gradually declines as more distant context is introduced. In contrast, the \texttt{item-churn} task on \texttt{amazon} exhibits a strong positive correlation with neighborhood size, consistently benefiting from a wider relational field.
These trends align with intuition: race completion tasks depend primarily on recent driver performance, while tasks such as item demand forecasting benefit from broader historical context.
Despite these task-level variations, \nametwo maintains strong performance across the full range of neighborhood sizes.
At the same time, subgraph breadth is a key factor for optimizing task-specific performance.

\paragraph{\nametwo remains robust under feature scarcity.}

We evaluate average model performance on a subset of RelBenchV1-classification under varying levels of feature scarcity, where missing values are randomly introduced into the context at increasing rates (\emph{cf.}~Fig.~\ref{fig:featdrop}).
We compare against XGBoost
(supervised) and TabICLv2 (in-context) baselines.
Tabular models are highly sensitive to this perturbation, with performance degrading rapidly as sparsity increases.
In contrast, \nametwo exhibits significantly stronger resilience.
Due to its relational structure, the model can access multiple alternative paths to relevant signals across tables, reducing its reliance on any single feature.

\paragraph{\nametwo maintains performance under link scarcity and cold-start settings.}

We evaluate average model performance
on a subset of RelBenchV1-classification under varying levels of link scarcity, where foreign key relationships are randomly removed at increasing rates (\emph{cf.}~Fig.~\ref{fig:edgedrop}).
This setup effectively models cold-start scenarios with limited or no historical interactions.
Notably, \nametwo remains highly robust under such structural degradation, maintaining strong performance even when up to 75\% of edges are removed.
In the extreme case, as link scarcity approaches 100\%, the model reduces to a single-table setting, where its performance aligns with that of recent tabular foundation models such as TabICLv2.

\paragraph{\nametwo is highly robust to noisy and uninformative features.}

We evaluate average model performance
on a subset of RelBenchV1-classification under varying levels of noise, where a number of noisy columns are injected into the context at increasing rates (\emph{cf.}~Fig.~\ref{fig:noise}).
While \nameone exhibits a steady decline in performance as noise increases, \nametwo remains significantly more resilient, with performance remaining nearly constant even under heavy noise injection.
We attribute this behavior to the task-conditioned intra-table processing of \nametwo, which enables the model to filter out irrelevant information early in the pipeline.

\subsection{Fine-Tuning \nametwo}
\label{subsec:finetuning}

Finally, we evaluate the fine-tuning capabilities of \nametwo to scale beyond current context size limits and leverage the full training set.
We use the SALT benchmark, where 100\% of labels are available, compared to only 0.5\% in the in-context setting.
We compare against the strongest baselines, DS+AutuGluon and DS+Carte, as well as the \nametwo in-context base model.

Results are reported in Table~\ref{tab:salttune} using Mean Reciprocal Rank (MRR).
Fine-tuning \nametwo yields substantial performance gains, improving MRR by 10 points over the best supervised baseline and by 6 points over the \name base model.
The fine-tuned model consistently outperforms all baselines across tasks.
Gains are particularly pronounced on the ``Sales Group'' and ``Payment Terms'' tasks, which denotes the tasks with the largest number of classes.
These results demonstrate that fine-tuning effectively overcomes the challenges of in-context learning in large-scale settings.
Notably, fine-tuning completes in under two minutes per task.

\begin{table}
\caption{\textbf{Test results on SALT when fine-tuning \nametwo.} Higher is better (MRR). \textcolor{green}{\raisebox{-0.2ex}{\rule{0.8em}{0.8em}}}~denotes the best, and \textcolor{green!50!white}{\raisebox{-0.2ex}{\rule{0.8em}{0.8em}}}~the second-best among all models and tasks.}
\label{tab:salttune}
\begin{minipage}{0.6\linewidth}
\centering
\setlength{\tabcolsep}{3pt}
\begin{tabular}{lcccc}
\toprule
\multirow{2}{*}{\textbf{Task}} & \href{https://arxiv.org/abs/2501.03413}{DS+Auto} & \href{https://arxiv.org/abs/2501.03413}{DS+} & \multicolumn{2}{c}{\textbf{\nametwo}} \\
& \href{https://arxiv.org/abs/2501.03413}{Gluon} & \href{https://arxiv.org/abs/2501.03413}{CARTE} & \multicolumn{1}{c}{ICL} & \multicolumn{1}{c}{Tuned} \\
\midrule
\small Plant & \cellsecond 0.99 & \cellsecond 0.99 & \cellsecond 0.99 & \cellfirst 1.00 \\
\small Shipping Point & \cellsecond 0.98 & 0.97 & \cellsecond 0.98  & \cellfirst 0.99 \\
\small Item Incoterm & \cellsecond 0.78 & 0.75 &  \cellsecond  0.78 & \cellfirst0.82 \\
\small Header Incoterm & 0.78 & 0.77 &  \cellsecond 0.81 & \cellfirst0.88 \\
\small Sales Office & 0.99 & 0.99 &  \cellfirst 1.00 &  \cellfirst 1.00 \\
\small Sales Group & 0.34 & 0.46 & \cellsecond 0.61 & \cellfirst0.74 \\
\small Payment Terms & 0.52 & 0.62 & \cellsecond 0.68 & \cellfirst0.86 \\
\small Shipping Condition & 0.74 & 0.74 & \cellsecond 0.79 & \cellfirst0.84 \\
\midrule
\textbf{Average} ($\uparrow$) & 0.77 & 0.79 & \cellsecond \textbf{0.83} & \cellfirst \textbf{0.89} \\
\bottomrule
\end{tabular}
\end{minipage}
\hfill
\begin{minipage}{0.39\linewidth}
\centering
\includegraphics[width=0.79\linewidth]{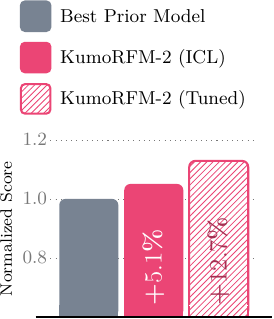}
\end{minipage}
\end{table}

\section{\name in Practice}

\nametwo is available through our SDK\footnote{Documentation: \url{https://kumo.ai/docs}} from Python 3.10 onwards via \texttt{pip install kumoai}.
It exposes a unified interface for graph construction, task specification, and model inference.
In addition to direct use through the Python SDK, the same abstractions are exposed as a collection of skills for modern agentic workflow integration.
Users can leverage in-context learning through our public inference endpoint, while fine-tuning capabilities are available via enterprise deployments.

\begin{figure}[t]
\begin{lstlisting}[style=python, language=Python,caption={\textbf{Using KumoRFM on Snowflake}: From data to predictions in a few lines of code.},label={lst:code}]
from kumoai import (*@\textbf{rfm}@*)

# Connect to a data warehouse and register a graph from its tables ######
graph = (*@\textbf{rfm.Graph.from}@*)_(*@\textbf{snowflake}@*)(
    connection=rfm.backend.snow.connect(...),
    tables=["users", "orders", "items"],
)

# Optionally refine the inferred schema #################################
# Set primary keys, semantic types, links, or derived columns:
graph["users"].primary_key = "user_id"
graph["items"]["category"].stype = "categorical"
graph.link(src_table="orders", fkey="user_id", dst_table="users")
graph["orders"].add_column(name="total", expr="price * quantity")

# Initialize KumoRFM. Any data compute is pushed down to the database ###
model = (*@\textbf{rfm.KumoRFM}@*)(graph)

# 1) Run ad-hoc predictions using Kumo's predictive query language ######
result = (*@\textbf{model.predict}@*)(
    # Predict if a user will place zero orders in the next 30 days:
    "PREDICT COUNT(orders.*, 0, 30, days)=0 FOR EACH users.user_id",
    indices=[0, 1, 2],         # User IDs to score
    anchor_time="2026-04-07",  # Reference timestamp for prediction
    explain=True,              # Whether to explain the prediction
    return_embeddings=True,    # Whether to return embeddings
    run_mode="fast",           # Context budget: fast|normal|best
    num_neighbors=[32, 32],    # Sampled subgraph width and depth
    lag_timesteps=10,          # Lagged targets per prediction example
    inference_config=dict(     # Ensemble/inference settings:
        num_estimators=4, column_shuffle=True, class_shuffle=True
    ),
)

# 2) Run predictions from explicit context and prediction tables ########
task = (*@\textbf{rfm.TaskTable}@*)(
    task_type="regression",
    context_df=pd.DataFrame({    # Examples used as context
        "user_id": [...], "time": [...], "target": [...]
    }),
    pred_df=pd.DataFrame({       # Examples to score
        "user_id": [...], "time": [...]
    }),
    entity_table_name="users",   # Entity table to predict for
    entity_column="user_id",     # Entity identifier column
    time_column="time",          # Context/prediction timestamp column
    target_column="target",      # Supervision/target column
)
result = (*@\textbf{model.predict}@*)_(*@\textbf{task}@*)(task)
\end{lstlisting}
\end{figure}

A typical workflow is illustrated in Example~\ref{lst:code}.
The SDK constructs a relational graph abstraction directly from database schema information.
For example, \name infers semantic structure such as primary keys, time columns, and inter-table links.
This automatic graph construction minimizes manual schema engineering while remaining fully overrideable through a declarative interface.

Once a graph is defined, users can instantiate \nametwo and issue predictions using a high-level predictive query language.
This interface allows users to express complex relational and temporal prediction tasks without explicitly materializing training datasets.
Alternatively, users can provide explicit context and prediction tables via a \texttt{TaskTable} abstraction, enabling tighter control over context examples.
Key runtime parameters allow users to trade off performance and efficiency, including \texttt{run\_mode} (context budget and model depth), \texttt{num\_neighbors} (subgraph sampling breadth and depth), \texttt{anchor\_time} (prediction timestamp), and \texttt{inference\_config} (ensemble configuration).

In contrast to traditional machine learning pipelines, \name does not require an explicit separation between training and prediction data.
Instead, context examples and prediction examples refer to the \emph{same} underlying graph.
For practitioners, this changes the primary abstraction from ``building and maintaining data pipelines for model training'' to ``specifying prediction tasks over a single relational database'' in which feature construction and context selection are handled automatically.

\paragraph{Best Practices.}

\name is designed to fully leverage relational and temporal structure, and excels in settings where rich interactions and histories are available.
In particular, datasets with meaningful entity relationships and temporal dynamics (\emph{e.g.}, user behavior over time) naturally unlock its full potential.
To get started, we recommend starting with a focused predictive query over a small set of relevant tables, and progressively expanding the graph as needed.
This iterative approach enables rapid prototyping while maintaining clarity over the signal captured by the model.
Care should be taken to ensure that relevant information is indeed reachable within a bounded number of hops, which can be controlled via \texttt{num\_neighbors}.
Once a query is stable, advanced parameters such as \texttt{run\_mode} and \texttt{inference\_config} can be tuned to optimize performance.

\paragraph{Resources.}
We provide a collection of notebooks\footnote{Notebooks: \url{https://github.com/kumo-ai/kumo-rfm}} to demonstrate common workflows, including:
\begin{itemize}[label=\textcolor{kumo1}{\textbullet},leftmargin=10pt]
\item \textbf{\href{https://colab.research.google.com/drive/17UppPgwzLrjkGtUpGjcxOYDcbp7koFqd}{Database Integration:}} Connecting to databases and constructing graphs from SQL tables
\item \textbf{\href{https://colab.research.google.com/github/kumo-ai/kumo-rfm/blob/master/notebooks/predictive_query.ipynb}{Predictive Query:}} Formulating predictive tasks using the Predictive Query Language
\item \textbf{\href{https://colab.research.google.com/github/kumo-ai/kumo-rfm/blob/master/notebooks/explanations.ipynb}{Explanations:}} Understanding and interpreting the different explanation methods in \name
\item \textbf{\href{https://github.com/kumo-ai/kumo-rfm/blob/master/notebooks/ecom_agent.ipynb}{MCP:}} Using the \name MCP server to get started with LLM-based workflows
\item \textbf{\href{https://colab.research.google.com/github/kumo-ai/kumo-rfm/blob/master/notebooks/salt.ipynb}{Evaluation:}} Evaluating \name on benchmark datasets and analyzing its performance
\item \textbf{\href{https://colab.research.google.com/github/kumo-ai/kumo-rfm/blob/master/notebooks/single_table.ipynb}{Single Tabular Data:}} Applying \name in non-relational settings on single tabular data
\end{itemize}

\section{Conclusion}
\label{sec:conclusion}

We presented \nametwo, an improved relational foundation model to provide training-free, ad-hoc predictions on relational databases, able to scale to production environments with 500B+ rows.
Our results indicate that the relational domain is reaching a critical inflection point.
A single pre-trained model can now rival or exceed the performance of task-specific pipelines.
This suggests a future where foundation models serve as the standard starting point for predictive modeling in databases.

\section*{Acknowledgements}

We thank the entire Kumo team for their invaluable support.
Special thanks go to Josh Przybylko, Adi Wadaskar, Hema Raghavan, Manush Murali, Raja Rao DV, Keti Jovanovska, Zack Drach, Vanja Josifovski, Myungwhan Kim, Bla\v{z} Stojanovi\v{c}, Siyang Xie, Alan Krumholz, Effy Fang, Abdullah Al-chihabi, Disha Dubey, Devan Johnson, Joseph Oliveira, Angela Liu, Miha Pelko, Vedaant Jain, Sally Liu, Aleksandar S. Sokolovski, Alex Porter, and Ramona Bendias.

\bibliography{refs}
\bibliographystyle{nat}

\end{document}